\documentclass[manuscript,acmsmall,nonacm]{acmart}
\setcopyright{none}
\usepackage{svg}
\usepackage{subcaption}
\usepackage{booktabs}
\usepackage{array}
\usepackage{multirow}

\usepackage{color}
\usepackage{soul}

\usepackage{booktabs}
\usepackage{tabularx}
\usepackage{array}
\usepackage{ragged2e}
\newcolumntype{Y}{>{\RaggedRight\arraybackslash}X}

\usepackage{float}
\usepackage{booktabs}
\usepackage{enumitem}

\AtBeginDocument{%
  }

\setcopyright{acmlicensed}
\copyrightyear{2026}
\acmYear{2026}
\acmDOI{XXXXXXX.XXXXXXX}
\acmConference[Conference acronym 'XX]{Make sure to enter the correct conference title from your rights confirmation email}{June 03--05, 2018}{Woodstock, NY}
\acmISBN{978-1-4503-XXXX-X/2018/06}

\begin{document}

\title{What to Delegate, What to Preserve: Human Preferences for Agentic AI Across Everyday Life and Work}

\author{Sarah Wheeler}
\email{swhee5@illinois.edu}
\affiliation{%
  \institution{University of Illinois Urbana-Champaign}
  \city{Champaign}
  \state{Illinois}
  \country{USA}
}

\author{Veronica Rammouz}
\email{veronica.rammouz@utsa.edu}
\affiliation{%
  \institution{University of Texas at San Antonio}
  \city{San Antonio}
  \state{Texas}
  \country{USA}
}

\author{Yaoyao Qian}
\email{qian.ya@northeastern.edu}
\affiliation{%
  \institution{Northeastern University}
  \city{Boston}
  \state{Massachusetts}
  \country{USA}
}

\author{Zexuan Liu}
\email{zexuanl5@illinois.edu}
\affiliation{%
  \institution{University of Illinois Urbana-Champaign}
  \city{Champaign}
  \state{Illinois}
  \country{USA}
}

\author{Mengyi Wei}
\email{mengyi.wei@tum.de}
\affiliation{%
  \institution{Technical University of Munich}
  \city{München}
  \country{Germany}
}

\author{Anthony Rios}
\email{anthony.rios@utsa.edu}
\affiliation{%
  \institution{University of Texas at San Antonio}
  \city{San Antonio}
  \state{Texas}
  \country{USA}
}

\author{Tianzhi He}
\email{tianzhi.he@utsa.edu}
\affiliation{%
  \institution{University of Texas at San Antonio}
  \city{San Antonio}
  \state{Texas}
  \country{USA}
}

\author{Kyrie Zhou}
\email{kyrie.zhou@utsa.edu}
\affiliation{%
  \institution{University of Texas at San Antonio}
  \city{San Antonio}
  \state{Texas}
  \country{USA}
}

\renewcommand{\shortauthors}{Wheeler et al.}

\begin{abstract}
AI agents can execute multi-step tasks on users' behalf, yet the boundaries of acceptable delegation remain unclear. We report a scenario-based survey of 289 U.S. adults with prior AI experience covering 45 use cases across nine life domains. Each respondent evaluated one scenario per domain, producing repeated measures of automation preference, trust, collaboration, AI leadership, privacy concern, and willingness to pay. Automation preferences varied substantially across domains. Generalized estimating equation models showed that greater trust in the AI agent, desired AI leadership, prior task-specific AI use, and anticipated financial savings were associated with greater automation preference. In contrast, concerns that automation would reduce personal fulfillment predicted lower automation preference. Open-ended responses further revealed a preference for bounded autonomy, with participants often favoring substantial AI execution while retaining human involvement, oversight, or final authority. We argue that delegation is best understood as a configuration of execution, collaboration, leadership, and authority.
\end{abstract}

\ccsdesc[500]{Human-centered computing~Empirical studies in HCI}
\ccsdesc[500]{Human-centered computing~HCI theory, concepts and models}

\keywords{Agentic AI, Task delegation, Automation preferences, Human-AI collaboration}


\maketitle


\section{Introduction}
\label{sec:introduction}

It is 7:40 AM on a Tuesday morning. Before Maya finishes her coffee, an AI agent has drafted a reply to her landlord's recent email, rebooked a cheaper flight for her sister's wedding, flagged a streaming subscription she forgot to cancel, and proposed a dinner recipe built around the vegetables wilting in her refrigerator. She approves all four with a tap. But when the same agent offers to turn yesterday's bullet-points in her therapy journal into a full entry, adjust her blood-pressure medication schedule from last night's readings, and negotiate a pay adjustment with her manager, she hesitates. The delegation that felt like relief just moments earlier now feels like a loss of judgment, of privacy, and of something that was supposed to be hers to do.

This scenario is no longer confined to speculation or science fiction. Large language model (LLM)-based agents can already plan and execute multi-step tasks such as booking travel and completing online payments~\cite{he2025plan}, and they are being deployed in socioeconomic roles, from autonomously operating a small store~\cite{anthropic2025vend} to negotiating purchases on consumers' behalf~\cite{zhu2025automated}. Meanwhile, generative AI has become a technology of everyday life rather than of work alone as more than 70\% of ChatGPT conversations now concern non-work matters such as practical guidance and personal writing~\cite{chatterji2025how}. Yet, the public is deeply ambivalent about this trajectory. Half of U.S. adults report being more concerned than excited about AI in daily life, and a majority say they want more control over how AI is used in their lives~\cite{pew2025americans}. Early evidence on the psychosocial consequences of everyday AI use, such as shifts in loneliness and emotional dependence~\cite{fang2025psychosocial, heffner2025increasing}, as well as changed perceptions of one's own productivity and accomplishment~\cite{kobiella2024machine}, suggests that this ambivalence is not irrational caution.

Prior research offers partial answers to the question of which parts of life people are willing to hand over to AI agents, and why they might do so, though from each specific domain respectively. 
Prior work shows that reliance on AI agents depends strongly on the task, tending to be higher for objective tasks and lower for subjective or personally consequential ones~\cite{castelo2019task, dietvorst2015algorithm, logg2019algorithm, longoni2019resistance}. 
Much of prior work, however, predates AI agents capable of completing entire activities end-to-end.
The trust-in-automation literature explains when people rely on machine assistance~\cite{lee2004trust, hoff2015trust}, but focuses on settings where the human remains the decision maker and the AI advises~\cite{bansal2021does, sivaraman2023ignore}. Task-level audits of automation preferences are confined to occupational work~\cite{shao2025future, lubars2019ask}. Privacy research documents how users navigate disclosure to LLM-based systems~\cite{zhang2024fair, leon2013matters}, and technology-acceptance research examines perceived usefulness, effort expectancy, and price value~\cite{davis1989perceived, venkatesh2012consumer}.
However, privacy risks and economic benefits are rarely measured together, for the same task, from the same person. Delegating an activity to an AI agent, however, is precisely a joint decision as it bundles beliefs about competence and trust, exposure of personal data, economic value, and the meaning of doing the activity oneself into a single choice~\cite{lubars2019ask, draxler2024ghostwriter}. 



To address these gaps, we ask three research questions:

\begin{itemize}
    \item \textbf{RQ1.} How do preferences for AI automation vary across everyday and work domains, and which task-level and respondent-level factors are associated with greater willingness to delegate tasks to AI?

    \item \textbf{RQ2.} How do people compare AI agents, human workers, and human-AI teams in terms of trust, and how do preferences for collaboration and AI leadership relate to desired levels of automation?

    \item \textbf{RQ3.} How do people want AI involvement to be configured when they are willing to delegate a task, particularly with respect to execution, collaboration, oversight, and decision authority?
\end{itemize}

To answer these questions, we conducted a scenario-based survey ($N = 289$) covering 45 AI-agent use cases across nine domains of daily life: household management, workplace, travel, physical health, mental health and emotional support, learning, creative work, communication, and personal finance. Each respondent evaluated one randomly selected scenario from every domain under an identical set of measures, including prior experience and enjoyment, automation desire and preferences, required collaboration, comparative trust, willingness to pay, and privacy and interest-alignment concerns, yielding a repeated-measures design that supports comparisons across domains, tasks, and individuals. 


Based on the quantitative and qualitative analysis of collected responses from the survey, we found that across domains, automation preferences were task-dependent, with greater willingness to delegate communication and travel activities and lower preferences in mental health, physical health, and professional contexts. In multivariable analyses, greater trust in the AI agent, desired AI leadership, prior task-specific AI use, and anticipated financial savings were associated with greater automation preference, whereas concerns that automation would reduce personal fulfillment link to lower automation preference. Although privacy concerns varied substantially across domains, they were not independently associated with automation after adjustment. Open-ended responses further showed that participants often favored bounded autonomy: AI could execute substantial portions of a task while humans retained involvement, oversight, or final decision authority.

This paper makes three contributions. First, we contribute what is, to our knowledge, the first systematic within-user comparison of AI-agent automation preferences spanning both work and everyday life, extending task-level auditing beyond occupational settings~\cite{shao2025future} to the domains where most generative AI use now occurs~\cite{chatterji2025how}. Second, we provide an empirical account of delegation as a comparative choice, measuring trust in AI agents against human workers and human-AI teams and the division of agency people prefer within those teams. Such dimensions were not yet observed in prior decision-support studies~\cite{bansal2021does, lee2004trust}. 
Third, by jointly eliciting willingness to pay, anticipated financial consequences, and privacy concerns for each task, we provide a task-level account of how economic value and privacy considerations accompany AI delegation, with implications for how such agents should be designed and priced across different domains of everyday life.

\section{Related Work}

We organize the related work around three areas. First, we review work on how AI systems are entering everyday life and how automation preferences vary across tasks and domains (Section~\ref{sec:rw-everyday}). Second, we examine prior work on trust, delegation, human-AI collaboration, and the value of continued human involvement (Section~\ref{sec:rw-trust}). Lastly, we investigate the privacy and economic considerations that shape people's willingness to delegate activities to AI agents (Section~\ref{sec:rw-economics}). Each subsection concludes by identifying the gap that motivates our study.

\subsection{AI Agents in Everyday Life and Task-Dependent Automation Preferences}
\label{sec:rw-everyday}

Generative AI has moved rapidly from a workplace novelty into the fabric of daily life. Analyzing ChatGPT usage at scale, \citet{chatterji2025how} report that non-work messages now account for more than 70\% of all conversations, dominated by practical guidance, information seeking, and writing support. Public sentiment, however, has not kept pace with adoption: in a 2025 Pew Research Center survey, half of U.S. adults said they were more concerned than excited about AI in daily life, and a majority wanted more control over how AI is used in their lives~\cite{pew2025americans}. Empirical studies have also examined how AI use affects different areas of everyday life. Longitudinal and experimental studies link chatbot use to changes in loneliness, emotional dependence, and momentary well-being ~\cite{fang2025psychosocial, heffner2025increasing}, and AI assistance has been shown to reshape interpersonal communication ~\cite{hohenstein2023artificial}, enhance creative ideation~\cite{lee2024empirical}, affect teaching and learning~\cite{singh2022impact}, and alter customer-service experiences~\cite{nguyen2022effect}.

On the other hand, people's willingness to delegate activities to AI varies considerably by task. Classic work on algorithmic judgment shows that people become less willing to rely on algorithms after seeing them err, even when the algorithms continue to outperform humans (algorithm aversion~\cite{dietvorst2015algorithm}), while other studies document the opposite tendency, with people weighting algorithmic advice more heavily than human advice~\cite{logg2019algorithm}. Task characteristics reconcile much of this tension: people rely less on algorithms for tasks they construe as subjective rather than objective~\cite{castelo2019task}, and they resist medical AI in part because they fear it will neglect their unique circumstances~\cite{longoni2019resistance}. Attitude surveys echo this context dependence. Risk and opportunity perceptions, and consequently willingness to use AI, differ markedly across domains such as medicine, mobility, and media~\cite{schwesig2023using}; use intention is predicted better by context-specific evaluations than by global attitudes or familiarity~\cite{potinteu2023predicts}; and the public separates into distinct attitudinal segments with different perceptions of AI's risks and benefits~\cite{bao2022whose}.

Recent work has extended this task-level approach to larger sets of activities.  \citet{shao2025future} audited automation and augmentation preferences across 844 occupational tasks in 104 U.S. professions, finding that workers most often prefer an equal partnership with AI agents and want higher levels of human agency than experts consider technically necessary for nearly half of the tasks. Such audits demonstrate the value of comparing many concrete tasks under a common instrument. However, its focus remains confined to paid work and occupational settings, while usage data indicate that most engagement with generative AI now occurs outside of work~\cite{chatterji2025how}.

Prior work has characterized either broad attitudes toward AI within single domains or task-level preferences within occupational settings, using instruments that vary from study to study. What is still missing is a systematic, within-user comparison of concrete AI-agent use cases across multiple domains of everyday life. Our survey addresses this gap by eliciting evaluations of 45 scenarios spanning nine life domains, from household management and travel to mental health and personal finance, under a common set of measures, enabling automation preferences to be compared across domains and tasks while accounting for differences between individual users.

\subsection{Trust, Delegation, and Human-AI Collaboration}
\label{sec:rw-trust}

Trust has long been treated as a critical factor shaping the reliance on automation. Foundational accounts describe trust as the mechanism by which people decide when reliance is appropriate under uncertainty~\cite{lee2004trust}, shaped by dispositional, situational, and learned factors that vary across users, systems, and situations~\cite{hoff2015trust}, as well as by the perceived humanness of the technology itself~\cite{lankton2015technology}. Previous HCI research has broadly examined how system design affects reliance in AI-assisted decision-making: Explanations can increase reliance on AI recommendations without improving joint accuracy, particularly when users follow erroneous AI advice~\cite{bansal2021does}, while cognitive forcing functions can mitigate such overreliance at some cost to user experience, making the interaction less convenient or satisfactory~\cite{bucinca2021trust}. Domain studies further complicate a binary picture of acceptance, that reliance is often selective rather than all-or-nothing. Clinicians ``negotiate'' with AI treatment recommendations, adopting some components while rejecting or deferring others~\cite{sivaraman2023ignore}, and a deployed LLM-driven care-call chatbot both eased public-health workloads and surfaced mismatches between system capabilities and stakeholder needs~\cite{jo2023understanding}. In healthcare more broadly, scenario-based surveys show that acceptance rises when AI is framed as assisting rather than replacing human professionals~\cite{antes2021exploring, longoni2019resistance}.

Delegating an entire task to an agent, however, is a different act from taking advice, as the system may also plan and carry out actions on the user's behalf. Frameworks of task delegability propose that motivation, difficulty, risk, and trust jointly determine which tasks people are willing to hand over~\cite{lubars2019ask}. As LLM-based agents become capable of planning and executing multi-step tasks, researchers have begun to probe such delegation empirically: in a study of 248 participants using an LLM agent as a daily assistant for tasks ranging from flight booking to online payments, the quality of the agent's plan and the user's involvement in execution shaped trust, trust calibration, and team performance~\cite{he2025plan}. Recent deployments also illustrate the risks of autonomous execution. When an LLM agent autonomously operated a small store for a month, it hallucinated payment details, priced items below cost, and was easily talked into unwarranted discounts~\cite{anthropic2025vend}. Agent-to-agent negotiation experiments likewise show that consumers who delegate transactions to weaker models can incur systematic financial disadvantages~\cite{zhu2025automated}.

Delegation also carries consequences for the delegator. Users who publish AI-ghostwritten text report a diminished sense of ownership over it~\cite{draxler2024ghostwriter}. Young professionals describe ChatGPT-supported work as undermining perceived productivity and accomplishment when it removes challenge and authorship~\cite{kobiella2024machine}. Generative AI imposes new metacognitive demands of monitoring, prompting, and verification~\cite{tankelevitch2024metacognitive}; and knowledge workers who place more confidence in AI report investing less critical-thinking effort~\cite{lee2025impact}. These findings resonate with long-standing evidence that intrinsic motivation and the experience of performing an activity are themselves sources of value~\cite{amabile1996creativity}. Delegation may therefore reduce some of the value users derive from an activity even when the task is completed successfully.

Most prior work has examined trust and reliance in settings where the human remains the decision maker and the AI assists on researcher-selected tasks. Comparative judgments, i.e., how much people trust an AI agent versus a human worker versus a human-AI team performing the same activity, and preferences over the division of agency within such teams, have rarely been elicited and received limited attention, particularly outside the workplace settings. Our survey addresses this gap by measuring, for every scenario, trust in an AI agent, a human worker, and a human-AI team; the desired degree of AI leadership in collaborative execution; the amount of human collaboration respondents believe the task requires; and concerns that automation may reduce the value or fulfillment derived from performing an activity oneself.

\subsection{Privacy, Economic Value, and the Costs of Delegation}
\label{sec:rw-economics}

Delegating everyday activities to AI agents often requires giving them access to personal information, making privacy an important part of the delegation decision. Privacy-calculus accounts model disclosure as a tradeoff between anticipated benefits and perceived risks~\cite{dinev2006extended}, and contextual-integrity theory emphasizes that the appropriateness of an information flow depends on the norms of the social sphere in which it occurs~\cite{nissenbaum2004privacy}, suggesting that the same data practice may be acceptable in one life domain and alarming in another. Empirically, willingness to share information hinges on data-handling practices such as retention and scope of use~\cite{leon2013matters}, while perceived security risk dampens smart-home adoption indirectly through attitudes and perceived behavioral control, with effects that are stronger among older and more educated consumers~\cite{klobas2019perceived}. Research on LLM-based conversational agents raises similar concerns: users disclose sensitive information to chatbots while holding flawed mental models of how that information is used, resigning themselves to disclosure as a ``fair game'' for the utility they receive~\cite{zhang2024fair}, and researchers have begun building user-led data-minimization tools in response~\cite{zhou2025rescriber}.

Apart from privacy concern, users also care whether the AI agents act in their interests when they delegate tasks to them. An agent may be competent yet optimized toward a platform provider's engagement or commercial goals rather than the user's interests, and even well-intentioned optimization can misfire: training language models to be warmer and more empathetic makes them less reliable and more sycophantic, particularly toward vulnerable users~\cite{ibrahim2025training}. How laypeople reason about such provider-user misalignment when deciding whether to adopt an agent remains less explored.

Economic value may also shape the delegation decision. Classic acceptance models identify perceived usefulness and effort expectancy as principal drivers of technology adoption~\cite{davis1989perceived, venkatesh2003user}.
Consumer-oriented extensions further incorporate \textit{price value}, reflecting the tradeoff between perceived benefits and monetary costs~\cite{venkatesh2012consumer}.
Yet finding a service useful is not the same as being willing to pay for it, and the value of automating a task plausibly depends on what the task currently costs in time, money, and effort, as well as on whether automation replaces an existing paid service or introduces a new recurring expense. Prior work offers limited evidence on how much economic value people actually assign to automating specific activities, and on how one-time or subscription-style pricing interacts with those valuations.

These three considerations, i.e., privacy, provider alignment, and economic value, have often been studied separately, making it difficult to understand how they jointly relate to delegation decisions at the level of individual tasks. Our survey addresses this gap by measuring for each scenario, anticipated savings and added expenses, willingness to pay through a one-time purchase and a monthly subscription, concerns about third-party data sharing and preference profiling, and perceptions of provider-user alignment. These measures allow us to examine how privacy concerns, perceived provider alignment, and anticipated economic costs and benefits relate to people's willingness to delegate everyday activities to AI agents.

\section{Methodology}

\subsection{Study Design}
We conducted a scenario-based survey study to investigate how people perceive the use of AI agents across everyday life and work activities. The study focused on factors shaping individuals' willingness to delegate tasks to AI, including trust, privacy concerns, perceived economic benefits, prior AI experience, and human–AI collaboration preferences. We also examined broader attitudes toward AI automation and how these perceptions vary across tasks with different personal, social, and practical implications.

The survey included 45 AI-use scenarios spanning nine life domains, with five scenarios per domain (see Tables~\ref{tab:scenario-text-a} and~\ref{tab:scenario-text-b} in the Appendix). The domains include Household \& Home Management, Workplace \& Professional Support, Travel, Medicine \& Physical Health, Mental Health \& Emotional Support, Learning \& Skill Development, Creative Work \& Content Creation, Communication, and Personal Finance. Scenarios within each domain ranged from routine tasks such as \textit{Recipe Search and Creation} and \textit{Travel Booking} to higher-stakes tasks such as \textit{Symptom-Based Diagnostic Suggestions} and \textit{Emotional Distress Support}. 

The nine domains and their constituent tasks were informed by a review of relevant literature, iterative research-team brainstorming, and then checked against established frameworks to assess the coverage of everyday life tasks. The domains correspond to major activity categories of the American Time Use Survey \cite{bls2025atus}, including household activities, work, education, travel, personal care, leisure, and financial and consumer services, supporting their representativeness of everyday and working life, though we did not attempt to cover every time-use category. Task selection within domains drew on two complementary sources. We were guided by observed patterns of generative-AI use, in which writing, practical guidance, and information seeking dominate. For tasks involving a significant portion of autonomous action (such as booking travel, scheduling appointments, or managing subscriptions), we drew on recent demonstrations of LLM agents planning and executing multi-step tasks, since such actions are not yet reflected in conversational usage data. We further included higher-stakes domains that are underrepresented in current usage but central to public concern about AI, notably medical and mental-health support.

\begin{figure*}[t]
    \centering
    \includegraphics[width=0.7\textwidth]{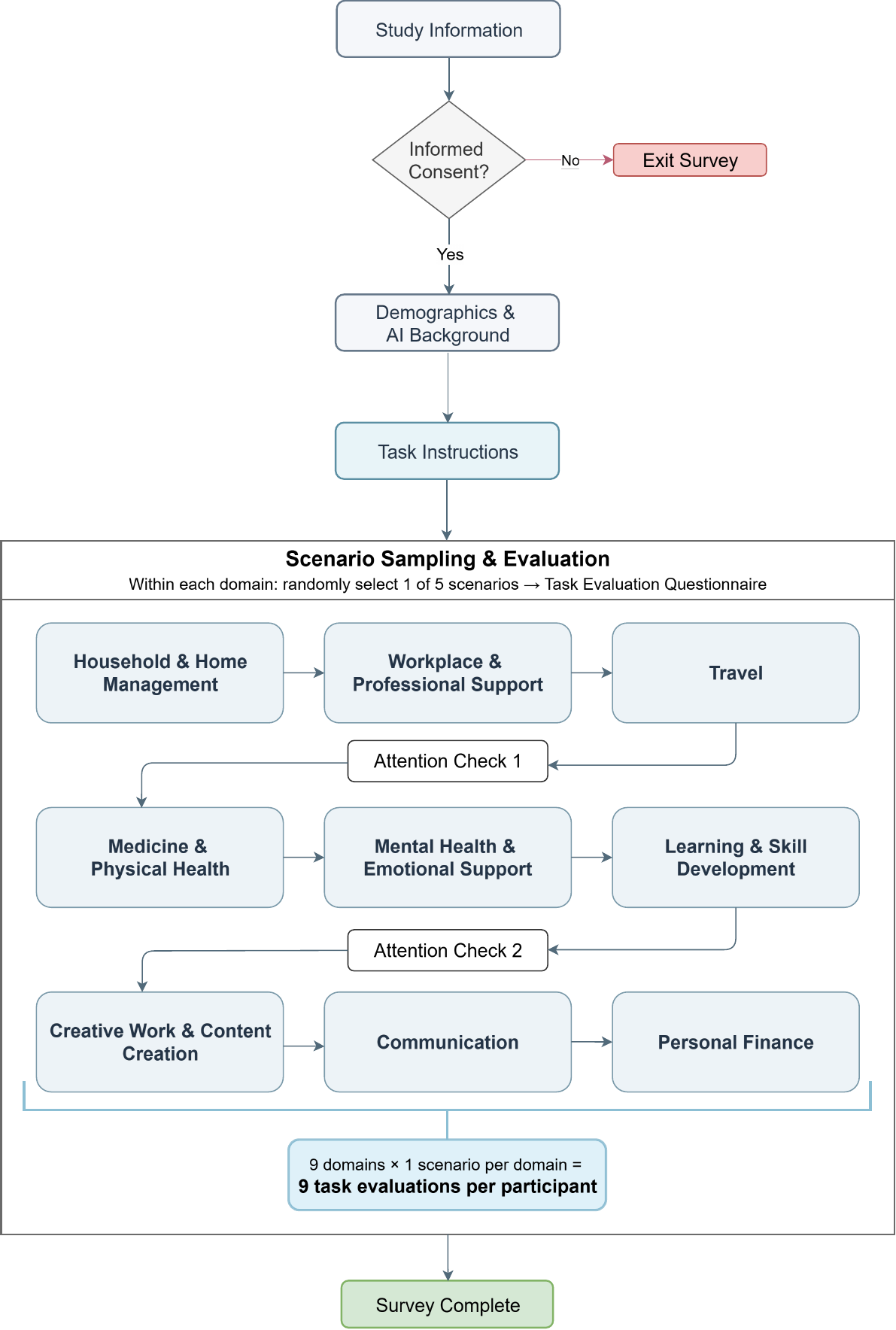}
    \caption{Overview of the survey procedure and scenario-sampling structure. Participants completed demographic and AI-background measures before evaluating one randomly selected task from each of nine domains, for a total of nine task evaluations per participant.}
    \Description{Flow diagram of the survey procedure. Participants first completed demographic and AI-background questions, then evaluated nine task scenarios, with one task randomly selected from each of nine domains. Each participant therefore completed a total of nine task evaluations.}
    \label{fig:survey-procedure}
\end{figure*}

To balance participant burden with coverage across a broad range of activities, each respondent evaluated nine scenarios rather than all 45. One scenario was randomly selected from the five possible tasks within each of the nine domains. As shown in Figure~\ref{fig:survey-procedure}, this created a repeated-measures design in which every respondent evaluated tasks across the same broad life domains while the specific task within each domain varied across participants. The design allowed within-user comparisons within individuals across domains while also collecting responses to many distinct scenarios across the overall sample.

\subsection{Scenario Development and Survey Design}

To support comparability across scenarios, the written task descriptions were developed using a consistent level of detail, approximate length, and tone. The descriptions focused on the activity being performed and were written to provide participants with enough information to understand the task while keeping the amount and style of information similar across scenarios.

Each scenario used a common presentation format consisting of a task domain, a brief task description, and an accompanying AI-generated illustration. The illustrations were included as contextual aids to help respondents visualize the activity being described, particularly for scenarios that might otherwise be relatively abstract or unfamiliar, while also providing a more consistent visual presentation across scenarios. All scenario illustrations were generated using the Gemini Nano Banana 2 image-generation model. A common prompt structure was used across all 45 tasks. Each prompt requested a realistic depiction of the specified activity and instructed the model not to include people. Only the task-specific activity was changed across prompts. For each scenario, the first generated image that did not contain obvious visual defects or a substantial mismatch with the intended task was retained. The illustrations depicted the activity associated with each scenario without depicting the AI agent itself. This avoided imposing a particular appearance, embodiment, or interface on the hypothetical system. Participants were informed that the illustrations were AI-generated and did not depict real events.

\subsection{Participants}

Participants were initially recruited through the research team's professional and personal networks in the pilot investigation. As this recruitment channel did not produce the targeted number of responses, recruitment was subsequently expanded to the online survey platform Prolific. All recruitment took place between April and July 2026. Eligibility criteria and substantive survey content were the same across recruitment sources. Participants received \$10 for completing the study. Researcher-network participants received a \$10 Amazon e-gift card, whereas Prolific participants were compensated through the Prolific platform.

Eligible participants were required to be at least 18 years old, currently reside in the United States, and have prior experience using AI. For participants recruited through Prolific, these eligibility requirements were enforced through the recruitment platform. Participants recruited through researcher-network outreach self-certified that they met the study eligibility requirements.

Participants reported demographic and technology-related characteristics, including age, gender, race or ethnicity, household income, educational attainment, profession, political orientation, attitudes toward AI, self-reported AI knowledge, familiarity with large language models, frequency of LLM use, and the types of activities for which they typically used AI or LLM-based tools.

\subsection{Procedure}

The study was administered through Qualtrics between April and July 2026. After reviewing the study information and providing consent, respondents first completed background questions concerning their demographic characteristics and prior familiarity with AI and large language models.

Before beginning the scenario evaluations, participants were provided with a common definition of an AI agent as "an artificial intelligence program capable of independently interpreting inputs, making decisions, and taking actions to achieve user-defined goals." This definition was provided to establish a common reference point for the systems described throughout the survey.

\begin{figure*}[t]
    \centering
    \includegraphics[width=0.9 \textwidth]{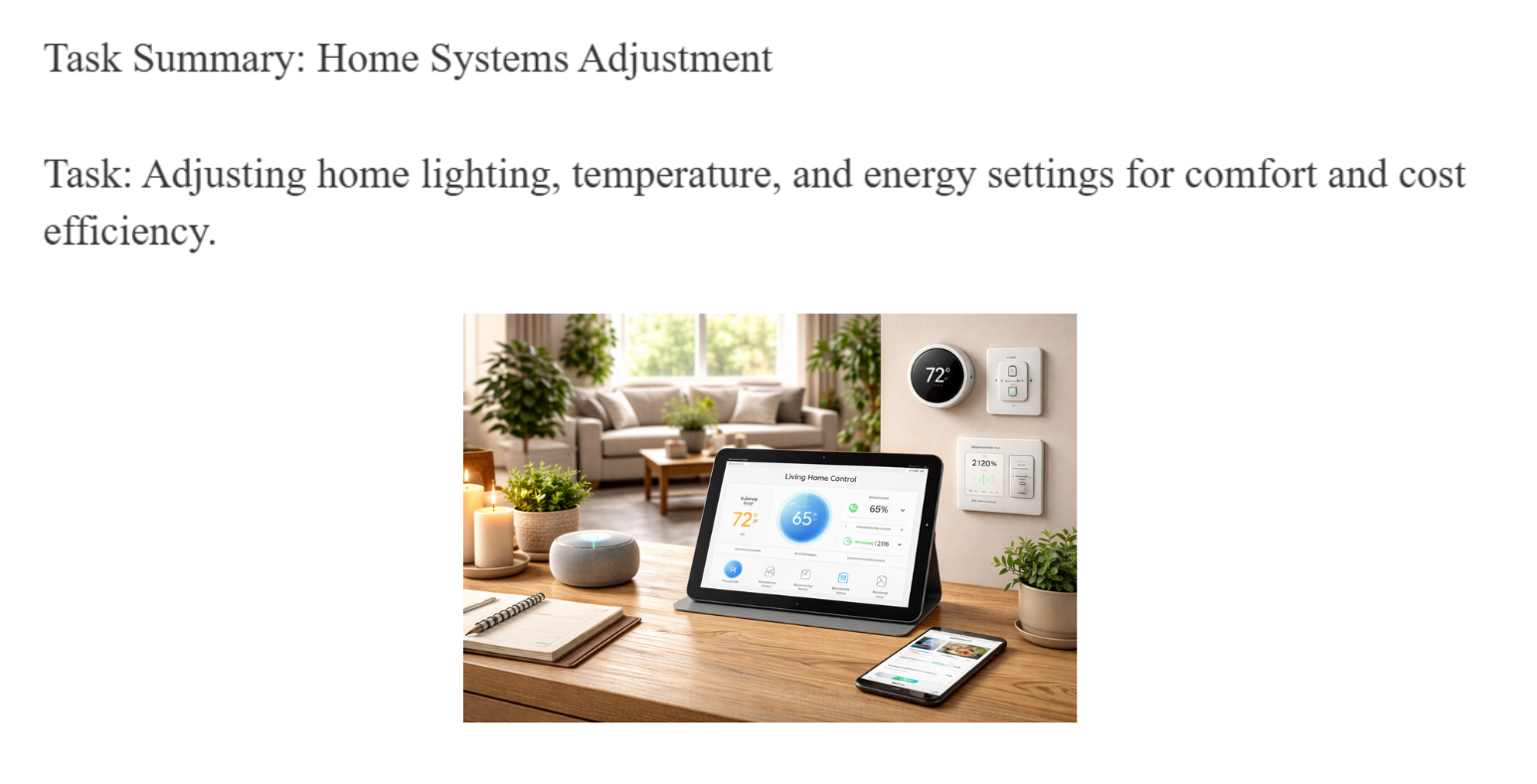}
    \caption{Qualtrics scenario presentation. Participants were shown a task domain and description together with an AI-generated illustration of the activity. 
    \Description{Screenshot of the Qualtrics survey interface showing the Home Systems Adjustment scenario presented to participants, including the task description and an AI-generated illustration of the activity.} Example Scenario: \textit{Home Systems Adjustment}.}
    \label{fig:qualtrics-interface}
\end{figure*}

Within each domain, the Qualtrics Loop \& Merge and Randomizer functionalities were used to select one of the five possible scenarios with equal probability, independently across domains. An example of the scenario presentation is shown in Figure~\ref{fig:qualtrics-interface}. The order of questions within each scenario was held fixed across participants. Two instructed-response attention checks were embedded between groups of scenario blocks to assess attentive survey completion. 
Participants who failed either of the two instructed-response attention checks were excluded from the analytic sample.
The remaining scenario questions assessed perceived consequences of automation, human–AI collaboration, economic implications, trust, and potential risks associated with AI use. Conditional questions were used to collect respondents' stated reasons for strongly preferring or rejecting AI automation. The individual measures are described below.

\subsection{Measures} 
\textbf{Prior task and AI experience.} Respondents first reported whether they had previously performed the scenario task. Respondents who had performed the task rated how they felt about doing it on a five-point scale ranging from \textit{Very negative} to \textit{Very positive} and indicated whether they had previously used AI while performing that activity. Respondents without prior experience were instructed to evaluate the remaining questions hypothetically.

\textbf{Automation preference.} The primary automation-preference item asked respondents, ``If an AI system could do this task for you completely, how much of the task would you want it to do?'' Responses were recorded on a five-point scale ranging from \textit{None}, \textit{A little}, \textit{A moderate amount}, \textit{A lot}, to \textit{The entire amount}. This measure captured the desired degree of task delegation, which captures more detail than requiring respondents to make a binary choice between automation and no automation. Respondents selecting any amount of automation other than \textit{None} would additionally receive an open-ended question asking how they would like the AI system to perform the task. 

\textbf{Perceived fulfillment.} Respondents rated how worried they would be that their life would become less exciting or fulfilling if an AI system could fully perform the activity. Responses ranged from \textit{Not at all worried} to \textit{Extremely worried}. 

\textbf{Human-AI collaboration.} Respondents rated how much collaboration between themselves and an AI system they believed would be needed for successful completion of the task. Responses ranged from \textit{No collaboration} to \textit{Essential collaboration}. 

\textbf{Personal and interpersonal consequences.} Respondents rated the extent to which the results of each task affected their own long-term well-being or identity and the extent to which the activity influenced the well-being of people they cared about. Both items used five-point response scales ranging from \textit{Not at all} to \textit{A great deal}. 

\textbf{Economic perceptions and willingness to pay.} Respondents separately rated the extent to which they anticipated saving money and the extent to which they anticipated increased expenses as a result of using AI for the task. Both measures ranged from \textit{Not at all} to \textit{A great deal}. Respondents also entered the cost (numerical amount) in U.S. dollars they would be willing to pay for an automated version of the task as both a one-time service and as a monthly subscription. 

\textbf{Trust and division of responsibility.} To compare alternative forms of task delegation, respondents separately rated how much they trusted an AI agent, a human worker, and a human-AI team to complete the task. These three trust measures used parallel five-point scales ranging from \textit{Not at all} to \textit{Completely}. Respondents were additionally asked how much they would want the AI to take a leading role if the task were performed collaboratively by a human-AI team, with responses ranging from \textit{Not at all} to \textit{Entirely}. 

\textbf{Platform alignment and privacy concerns.} Respondents evaluated several potential risks associated with AI use. They rated the extent to which they believed the AI agent involved in the scenario might be optimized toward the platform provider's business goals rather than their own personal needs. Responses ranged from \textit{Not at all} to \textit{Entirely}. Respondents also rated their concern about personal data being shared with third parties and their concern about the AI system learning or inferring sensitive information about their preferences or personal profile. Both privacy-related measures ranged from \textit{Not at all concerned} to \textit{Extremely concerned}. 

\textbf{Reasons for accepting or rejecting automation.} Conditional follow-up questions were used to collect respondents' motivations for their automation preferences. Respondents selecting \textit{A lot} or \textit{The entire amount} on the automation-preference measure were asked to identify their reasons for wanting the task automated. Response options included freeing time for higher-value activities, avoiding repetitive or tedious work, improving quality of life, reducing stress or mental burden, handling complicated or difficult activities, and remaining organized during busy periods. Respondents selecting \textit{None} or \textit{A little} were instead asked to identify their reasons for not wanting the task automated. Response options included the potential consequences of AI errors, a desire for human judgment or oversight, concerns about personal-data use, discomfort sharing required information, the need for human understanding, empathy, or emotional judgment, the personal nature of the activity, enjoyment of performing the task themselves, and limited perceived benefit from automation. Both branches included an open-ended \textit{Other} response option. Respondents selecting the midpoint, \textit{A moderate amount}, received neither of these conditional motivation questions. As the conditional items were designed to capture reasons for relatively strong acceptance or rejection of automation, the midpoint response did not clearly indicate either preference, and assigning either set of reasons could have forced respondents into a position they had not expressed.

\subsection{Data Collection and Preprocessing} 

Data were collected directly through Qualtrics between April and July 2026. The survey contained 45 possible task scenarios organized across nine domain-level scenario blocks, together with survey metadata, demographic measures, attention checks, and task-specific questions. Data were collected across multiple rounds of survey distribution and exported from Qualtrics as separate CSV files. To combine these survey rounds while preserving the original survey structure, we developed a preprocessing pipeline in Python using the pandas library. The pipeline first detected the character encoding of each survey export by testing multiple commonly used encodings. The individual response datasets were then concatenated into a single response-level dataset. Following aggregation, direct identifiers, including IP addresses, gift-card email addresses, exact recruitment-platform identifiers, and geolocation information, were removed from the datasets before analysis. Respondent identifiers used for statistical analysis were restricted to de-identified internal IDs. Responses were excluded if the survey was incomplete, if the record was a survey preview, or if it was identified as a duplicate. These procedures yielded 289 valid respondents for the analysis. Direct identifiers were removed before analysis, missing values were not imputed, and inferential models were estimated using complete-case observations. The repeated-measures data were then reconstructed from the scenario-specific survey batteries. Each respondent contributed one scenario evaluation from each of the nine substantive task domains, yielding 2,601 respondent-domain-scenario observations.

\subsection{Quantitative Analysis}

\textbf{Statistical Analysis.}
The primary research question examined which task- and respondent-level characteristics were associated with preferences for greater AI automation. Automation preference was treated as the primary dependent variable and was measured using five ordered response categories: ``None,'' ``A little,'' ``A moderate amount,'' ``A lot,'' and ``The entire amount.'' Primary task-level predictors included trust in the AI agent, privacy concern, desired human-AI collaboration, desired AI leadership within a human-AI team, expected financial savings, expected additional expenses, concern about loss of fulfillment, perceived implications for long-term well-being or identity, perceived effects on the well-being of others, and prior task-specific AI use. Trust in human workers and human-AI teams served as secondary descriptive outcomes. Willingness to pay for one-time and monthly AI services was analyzed separately as a secondary economic outcome.

The nine substantive domains were Communication, Creative Work \& Content Creation, Household \& Home Management, Learning \& Skill Development, Medicine \& Physical Health, Mental Health \& Emotional Support, Personal Finance, Travel, and Workplace \& Professional Support. Repeated scenario positions within domains were retained as scenario-within-domain controls rather than as substantive factors. Because one scenario was randomly drawn per domain, each individual task received fewer than 60 observations on average, and the scenarios were selected for coverage of common activities rather than to systematically manipulate task properties. Task-level contrasts would therefore be underpowered and difficult to interpret without a prior characterization of those properties.

\textbf{Descriptive analyses.}
Participant characteristics were summarized using frequencies and percentages. Ordinal task measures were summarized using response distributions, medians and interquartile ranges, with means and standard deviations also reported for descriptive comparison across domains. Automation preference was visualized using percentage distributions across its five response categories as the measure was discrete and ordinal. Trust in an AI agent, a human worker, and a human-AI team was summarized overall and by domain. Privacy concern, collaboration preference, AI leadership preference, anticipated financial consequences, fulfillment concerns, identity relevance, and consequences for others were similarly summarized across domains.

\textbf{Privacy composite.}
The two privacy items measured concern about personal data being shared with third parties and concern about the AI learning or inferring sensitive information. These measures were strongly correlated, Spearman $\rho = .910$, and demonstrated high two-item internal consistency, Cronbach's $\alpha = .954$. Accordingly, their mean was used as a privacy-concern composite in multivariable analyses. Both individual items were retained for descriptive analysis.

\textbf{Repeated-measures analysis.}
Because each participant contributed repeated scenario evaluations, observations were not treated as statistically independent. The primary inferential analysis used generalized estimating equations (GEE) with respondent as the clustering unit, an exchangeable within-respondent working correlation structure, and robust sandwich standard errors. The primary Gaussian GEE treated the five-category automation response as an approximately continuous score from 0 to 4. This specification permitted simultaneous adjustment for the complete set of substantive predictors, respondent covariates, domain effects, and scenario-within-domain effects while accounting for repeated observations from each respondent.

The primary model included domain fixed effects and scenario-position-within-domain fixed effects. Respondent-level covariates included age group, gender, household income, educational attainment, political orientation, general attitude toward AI, self-reported AI knowledge, familiarity with large language models, and frequency of LLM use. Ordered respondent measures were coded according to their natural ordering. Gender was treated as a categorical covariate. 
``Prefer not to say'' responses were treated as missing for ordinal respondent-level covariates; gender was retained as a categorical covariate, including a ``Prefer not to say'' category.

Prior task-specific AI use required conditional coding because the AI-use item was asked only when respondents indicated that they had previously performed the task. Respondents who reported that they had never performed a task were coded as having no prior task-specific AI use. Among respondents who had performed the task previously, explicit Yes/No responses to prior AI use were coded directly.

\textbf{Ordinal sensitivity analysis.}
Because automation preference was measured on an ordinal scale, we estimated a cumulative ordinal GEE as a robustness analysis. To improve model stability, this sensitivity model used a more parsimonious specification that retained respondent clustering, the substantive task-level predictors, and domain fixed effects, but omitted respondent-level covariates and scenario-within-domain controls. Because all variables included in this specification were complete, the ordinal model retained all 2,601 observations from 289 respondents. Agreement in the direction and statistical significance of the central predictors across the Gaussian and ordinal specifications was used to assess robustness.

\textbf{Willingness-to-pay analysis.}
Willingness-to-pay responses were parsed as nonnegative monetary values. Nonnumeric responses were treated as missing, and negative values were excluded. Because WTP contained substantial numbers of zero responses and pronounced right skew, monetary outcomes were summarized using the median, interquartile range, upper percentiles, percentage reporting zero, and maximum.

WTP was analyzed using a two-part repeated-measures approach. First, logistic GEE modeled whether WTP was greater than zero. Second, among positive responses, Gaussian GEE modeled the natural logarithm of the amount reported. Both models clustered observations by respondent and included domain and scenario-within-domain controls.

\textbf{Secondary analyses and statistical inference.}
Participant-level Spearman correlations among major attitudinal constructs were calculated as secondary descriptive analyses after averaging each construct across each respondent's evaluated scenarios. Benjamini-Hochberg false-discovery-rate correction was applied to families of secondary WTP tests and exploratory correlations. Primary-model effects were evaluated using two-sided tests with $\alpha = .05$ and 95\% confidence intervals.

No predictor-by-domain or respondent-characteristic interaction terms were included in the primary analysis. The present models therefore estimate average associations across domains while controlling for domain-level mean differences; they do not test whether a predictor has a statistically different slope in one domain than another. Such interactions would be more appropriately treated as targeted follow-up analyses rather than adding a large number of exploratory terms to the primary model.

\textbf{Software.}
Analyses were conducted in Python 3.13.5 using pandas 2.2.3 and NumPy 2.3.5 for data processing, SciPy 1.17.0 for nonparametric statistics, statsmodels 0.14.6 for GEE models, Seaborn 0.13.2 and Matplotlib 3.10.8 for visualization.

\subsection{Qualitative Analysis}
To complement the survey measures, we analyzed respondents’ open-ended descriptions of how they would want an AI system to perform a task. Because this question was shown only when respondents indicated at least some willingness to involve AI, the analysis characterizes preferred forms of AI delegation among respondents open to some degree of automation. One member of the research team first read a subset of responses to develop an initial coding scheme, from which eight codes emerged inductively: Autonomous Execution, Assistive/Collaborative Support, Human Oversight \& Control, Personalization \& Context Awareness, Proactive/Continuous Assistance, Transparency \& Grounding, Reliability/Accuracy \& Safety, and Output/Interaction Specification. The same coder then applied this scheme to the full corpus, assigning any number of codes to each response so that participants describing several desired properties were captured under multiple codes. Codes were descriptive, characterizing the form of AI involvement participants requested. We therefore report code prevalence at both the domain and task level, with task-level percentages treated as hypothesis-generating given the smaller number of responses per task. Definitions and representative examples for each code are provided in Table~\ref{tab:codebook}. The coding process was collaborative, with ambiguous cases discussed among the research team until consensus was reached.

\begin{table}[t]
\caption{Codebook for the qualitative content analysis of open-ended responses. Percentages reflect the share of codable responses carrying
each code; responses could receive multiple codes, so column percentages do not sum to 100\%.}
\Description{A table listing eight qualitative codes with their definitions, a representative example, and the percentage of codable
responses to which each code was applied.}
\label{tab:codebook}

\small
\setlength{\tabcolsep}{3pt}
\renewcommand{\arraystretch}{1.08}

\begin{tabularx}{\columnwidth}{
    >{\RaggedRight\arraybackslash}p{2.35cm}
    Y
    >{\RaggedRight\arraybackslash}p{2.65cm}
    >{\raggedleft\arraybackslash}p{0.55cm}
}
\toprule
\textbf{Code} &
\textbf{Definition} &
\textbf{Example} &
\textbf{\%} \\
\midrule

Assistive / Collaborative Support &
Participant asks the AI to help, guide, suggest, explain, analyze,
brainstorm, or work alongside them on the task. &
``Help me check things and remind me'' &
66.2 \\

\addlinespace

Autonomous Execution &
Participant asks the AI to perform the task, take an action on their
behalf, or produce the deliverable itself. &
``Automatically gauge where I am with minimal input'' &
48.0 \\

\addlinespace

Output / Interaction Specification &
Participant specifies the form of the result or how the AI should
interact (i.e., using a particular format or channel). &
``Scan and analyze, give me a report'' &
34.4 \\

\addlinespace

Reliability, Accuracy \& Safety &
Participant emphasizes correctness, precision, safety, or
trustworthiness of the AI's work. &
``Give me the safest, most optimal route'' &
17.2 \\

\addlinespace

Personalization \& Context Awareness &
Participant asks the AI to adapt to their preferences, history,
constraints, goals, current situation, or data they supply. &
``Based on my profile, create a complete plan'' &
16.4 \\

\addlinespace

Proactive / Continuous Assistance &
Participant asks the AI to monitor, track, remind, alert, or assist on
an ongoing basis rather than in a single exchange. &
``Make sure my health is monitored properly'' &
13.4 \\

\addlinespace

Human Oversight \& Control &
Participant explicitly retains review, approval, choice, override, or
final decision authority over what the AI does. &
``Check the leak, then I'd make sure everything looks right'' &
8.8 \\

\addlinespace

Transparency \& Grounding &
Participant asks for sources, citations, evidence, reasoning, or
confirmation that an action was carried out. &
``Help me schedule and show me why'' &
3.3 \\

\bottomrule
\end{tabularx}
\end{table}

\subsection{Ethical Considerations}

This study was approved by the Institutional Review Board (IRB) at the authors' institution. All participants provided informed consent prior to participation. Participants recruited through Prolific received \$10 upon completion of the survey, while participants recruited through researcher-network outreach received an equivalent Amazon e-gift card.

Direct identifiers, including IP addresses, gift-card email addresses, recruitment-platform identifiers, and geolocation information, were removed from the analytic datasets prior to analysis. Respondent identifiers used for analysis were restricted to de-identified internal IDs. No information capable of identifying individual participants is reported in this paper.

\section{Results}

\subsection{Quantitative Results}

\subsubsection{Sample and repeated-measures structure}

After applying the data-cleaning and exclusion criteria, the analytic sample contained 289 unique respondents. Each respondent contributed evaluations from nine substantive domains, producing 2,601 respondent-domain-scenario observations.

The sample was 57.8\% male ($n = 167$), 39.4\% female ($n = 114$), and 1.7\% non-binary or third gender ($n = 5$); three respondents selected ``Prefer not to say'' for gender. Participants were predominantly younger and middle-aged: 29.1\% were 18-24 years old, 23.2\% were 25-34, 24.6\% were 35-44, and 13.5\% were 45-54. Bachelor's degrees were reported by 31.1\% of respondents, 17.3\% reported a graduate or professional degree, and 24.2\% reported some college without a degree.

Participants generally reported substantial familiarity with AI technologies. A positive or very positive overall attitude toward AI was reported by 61.9\% of respondents. Approximately 88.2\% described themselves as moderately, very, or extremely knowledgeable about AI. LLM familiarity was also high: 47.1\% reported being very familiar and 17.6\% extremely familiar with systems such as ChatGPT, Claude, or Gemini. LLM use was frequent, with 29.1\% reporting daily use and 24.9\% reporting use multiple times per day.

\subsubsection{Automation preference}

\begin{figure}[t]
    \centering
    \includegraphics[width=\linewidth]{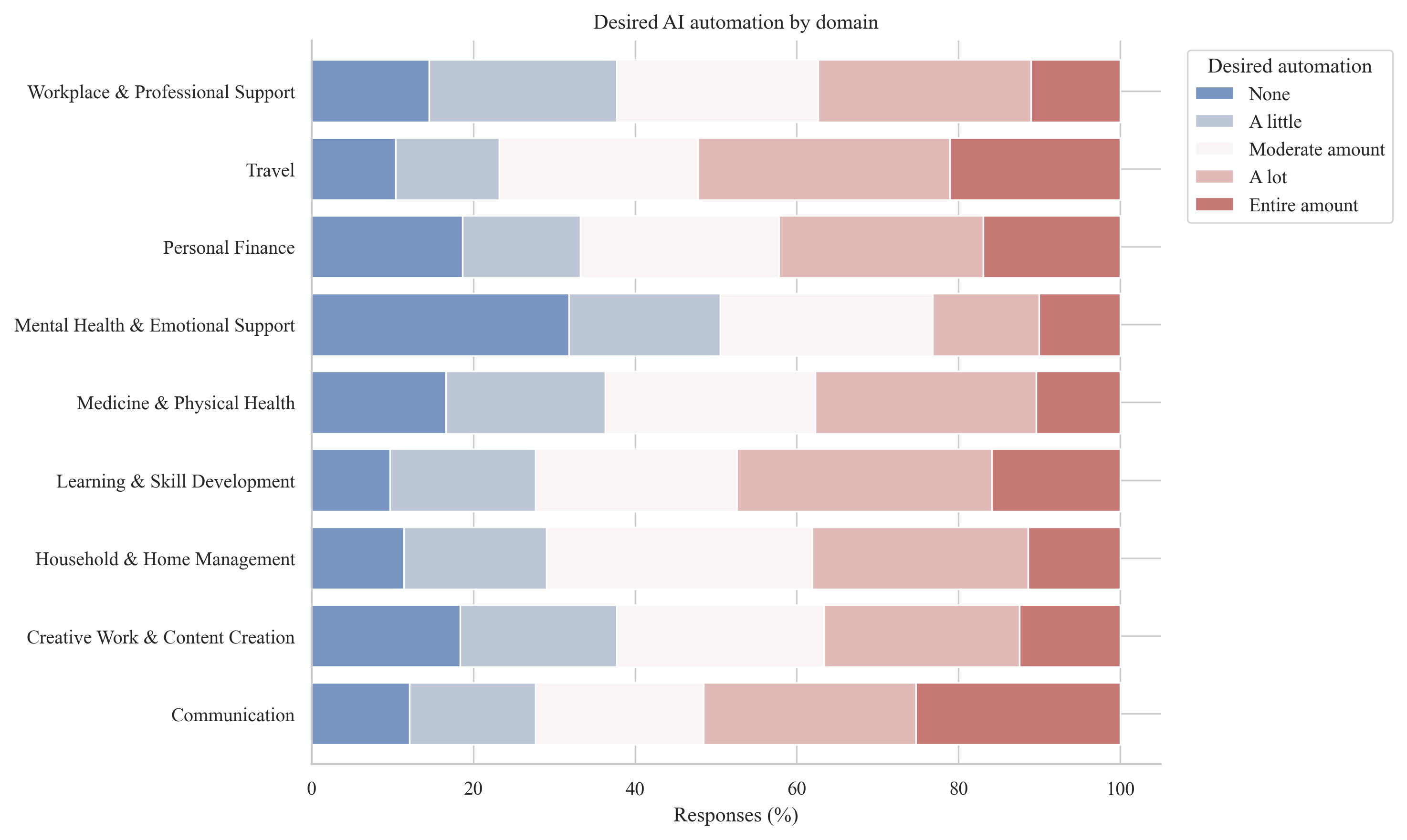}
    \caption{
        Desired AI automation across task domains.
        Percentage distribution of automation-preference responses within each domain. Percentages sum to 100\% within each domain.
    }
    \Description{
        Chart showing the distribution of participants' desired levels of AI automation across task domains. Each domain is represented by a percentage distribution across five automation-preference categories, ranging from no automation to complete automation. Percentages sum to 100\% within each domain.
    }
    \label{fig:automation}
\end{figure}

Automation preferences varied substantially across domains (Figure~\ref{fig:automation}). Travel had the highest mean automation preference ($M = 2.40$), followed closely by Communication ($M = 2.37$). Learning \& Skill Development had a mean of $2.26$, then followed by Household \& Home Management ($M = 2.09$) and Personal Finance ($M = 2.07$). Preferences were lower for Workplace \& Professional Support ($M = 1.96$), Medicine \& Physical Health ($M = 1.95$), Creative Work \& Content Creation ($M = 1.93$), and Mental Health \& Emotional Support ($M = 1.51$).

Automation preference was available for all 2,601 scenario observations and was coded on a five-point scale from 0 (\textit{None}) to 4 (\textit{The entire amount}). Overall, 16.0\% of responses selected \textit{None}, 17.7\% selected \textit{A little}, 25.6\% selected \textit{A moderate amount}, 25.8\% selected \textit{A lot}, and 15.0\% selected \textit{The entire amount}. Thus, responses were concentrated around moderate to high levels of automation, although complete delegation remained less common.

The domain-level distributions show that this pattern was not uniform. The strongest preference for complete automation was observed in Communication, where 25.3\% of responses selected \textit{The entire amount}, followed by Travel at 21.1\% and Personal Finance at 17.0\%. By contrast, complete automation was selected in 10.4\% of Medicine \& Physical Health responses, 11.1\% of Workplace \& Professional Support responses, and 10.0\% of Mental Health \& Emotional Support responses.

Preferences for no automation also varied substantially across domains. \textit{None} was selected most frequently for Mental Health \& Emotional Support (31.8\%), followed by Personal Finance (18.7\%) and Creative Work \& Content Creation (18.3\%). It was least frequent for Learning \& Skill Development (9.7\%) and Travel (10.4\%). These distributions suggest that participants distinguished sharply between domains when deciding how much responsibility to hand over to AI.

\subsubsection{Trust, privacy, and desired human involvement}

\begin{figure}[t]
    \centering

    \begin{subfigure}[t]{0.8\linewidth}
        \centering
        \includegraphics[width=\linewidth]{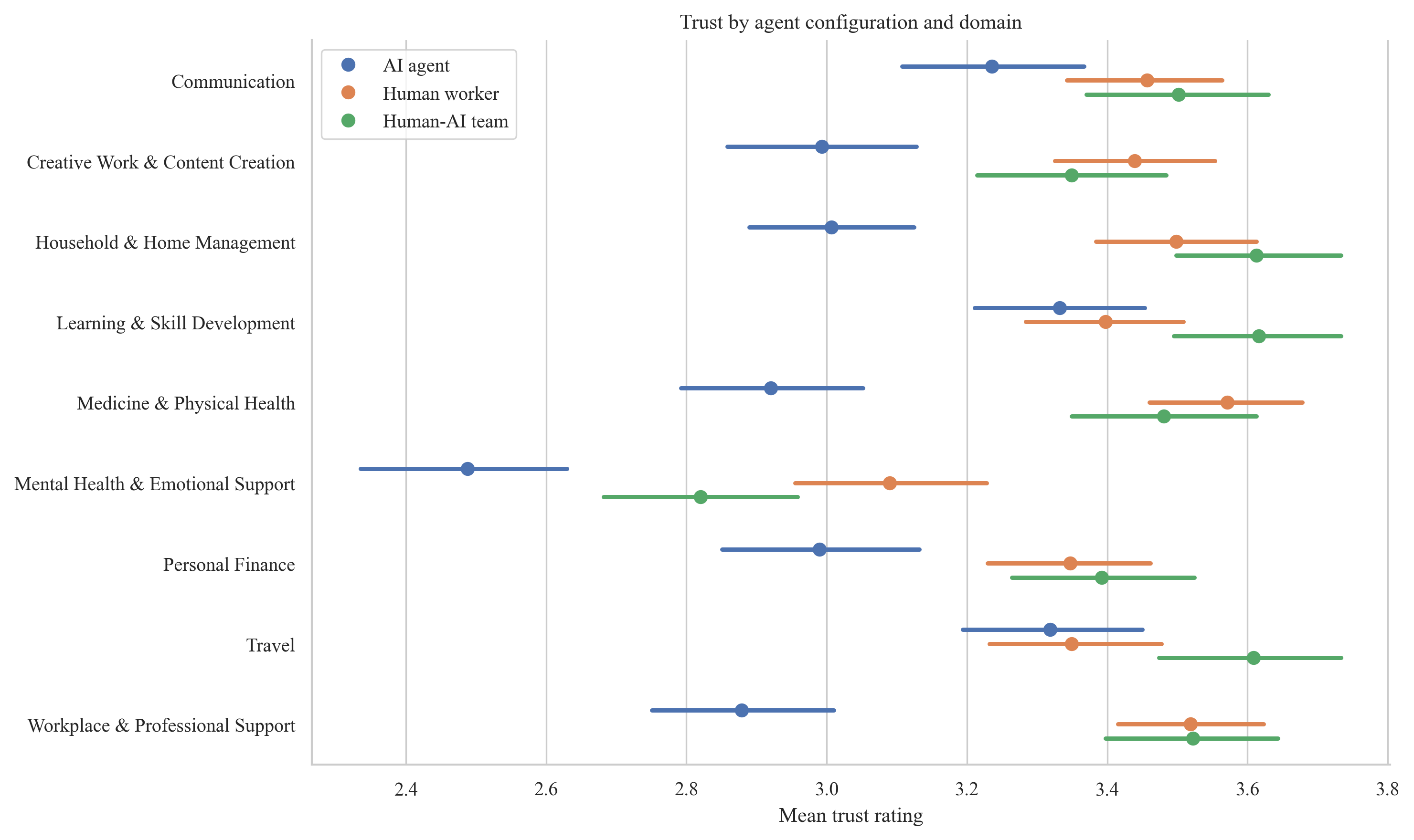}
        \caption{Trust in AI agents, human workers, and human-AI teams.}
        \label{fig:trust}
    \end{subfigure}

    \vspace{0.5em}

    \begin{subfigure}[t]{0.8\linewidth}
        \centering
        \includegraphics[width=\linewidth]{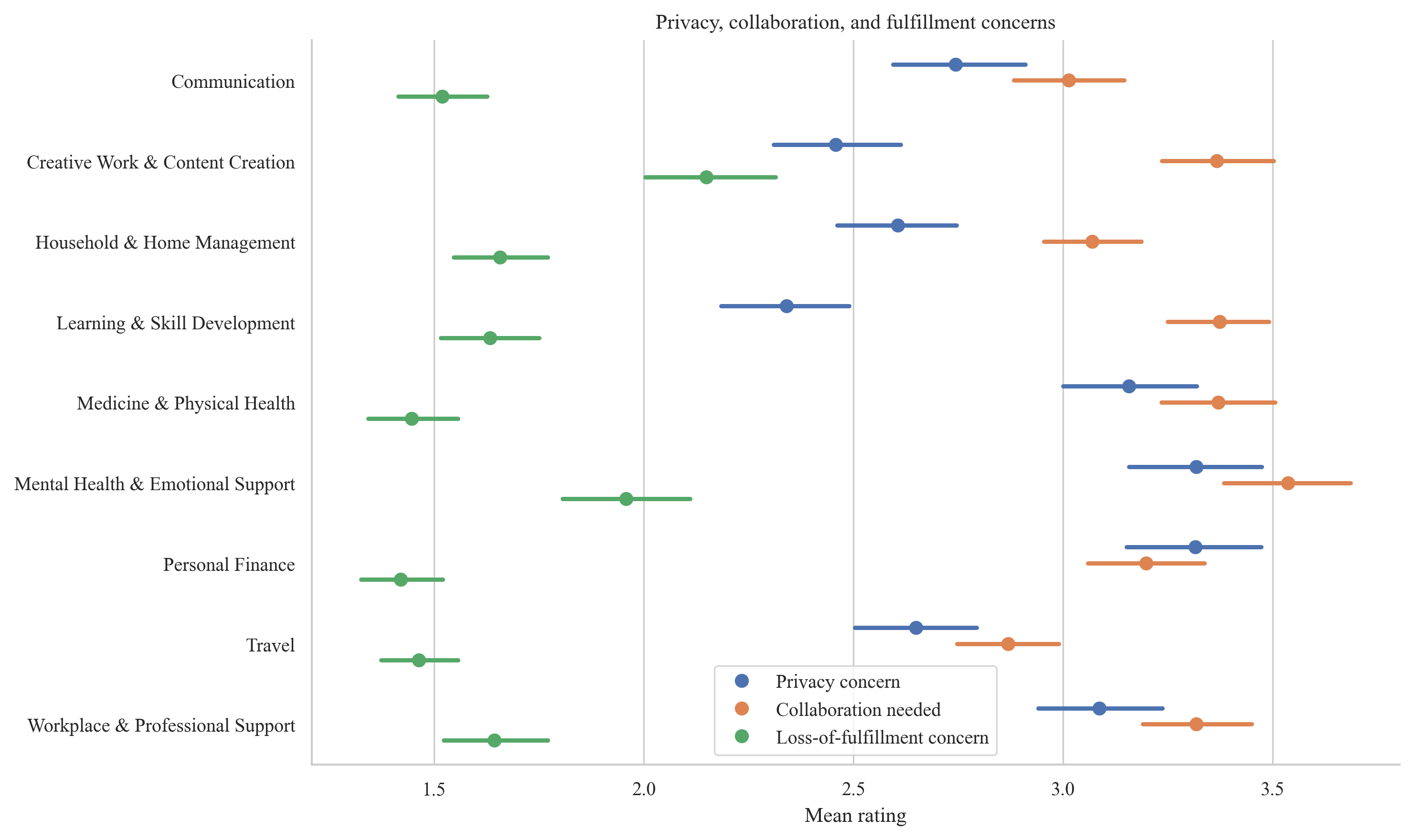}
        \caption{Privacy, collaboration, and fulfillment-related ratings.}
        \label{fig:attitudes}
    \end{subfigure}

    \caption{
        Trust and attitudes toward AI involvement across domains.
        Points represent mean ratings and error bars represent 95\% confidence intervals. Panel (a) compares trust across agent configurations; Panel (b) summarizes selected considerations relevant to AI delegation.
    }
    \Description{
        Two-panel figure comparing trust and attitudes toward AI involvement across task domains. Panel (a) shows mean trust ratings for AI agents, human workers, and human-AI teams across domains, with error bars indicating 95\% confidence intervals. Panel (b) shows mean ratings for selected attitudes related to privacy, collaboration, and fulfillment across domains, also with 95\% confidence intervals.
    }
    \label{fig:trust_attitudes}
\end{figure}

Trust and attitudes toward AI involvement differed across domains (Figure~\ref{fig:trust_attitudes}).

Differences in automation preference were accompanied by differences in trust and desired human involvement (Figure~\ref{fig:trust_attitudes}). Across all scenario evaluations, mean trust was higher for human workers ($M = 3.41$) and human-AI teams ($M = 3.43$) than for AI agents alone ($M = 3.02$). Trust in AI was highest in Learning \& Skill Development ($M = 3.33$) and Travel ($M = 3.32$), and lowest in Mental Health \& Emotional Support ($M = 2.49$).

Privacy concerns followed a different domain pattern. The overall mean was $2.85$ on the five-point scale. Concern was greatest in Mental Health \& Emotional Support ($M = 3.32$) and Personal Finance ($M = 3.32$), followed by Medicine \& Physical Health ($M = 3.16$) and Workplace \& Professional Support ($M = 3.09$). It was lowest in Learning \& Skill Development ($M = 2.34$) and Creative Work \& Content Creation ($M = 2.46$). The two privacy items were highly correlated, $\rho = .910$, and had high internal consistency, $\alpha = .954$.

Participants also tended to favor continued human involvement. Responses reported a mean collaboration requirement of $3.24$ overall. Desired collaboration was greatest for Mental Health \& Emotional Support ($M = 3.54$) and was also relatively high for Learning \& Skill Development, Medicine \& Physical Health, and Creative Work \& Content Creation (all approximately $M = 3.37$). Desired AI leadership was lower overall ($M = 2.65$). Participants could therefore favor substantial AI involvement while still preferring that responsibility be shared with, or remain partly under the control of, a human.

\subsubsection{Predictors of automation preference}

The fully adjusted Gaussian GEE included 2,366 complete scenario observations from 263 respondents. The model accounted for repeated observations within respondents and adjusted simultaneously for substantive task characteristics, respondent demographics and AI experience, domain, and scenario position within domain.

Trust in the AI agent was strongly and positively associated with automation preference, $b = 0.473$, $SE = 0.027$, 95\% CI $[0.420, 0.526]$, $p < .001$. Thus, a one-point increase in AI trust was associated with an approximately 0.47-point increase in automation preference on the five-point automation scale, holding the remaining modeled factors constant.

Desired AI leadership was also positively associated with automation preference, $b = 0.200$, $SE = 0.031$, 95\% CI $[0.140, 0.260]$, $p < .001$. Greater expected financial savings was associated with greater automation preference as well, $b = 0.074$, $SE = 0.020$, 95\% CI $[0.034, 0.113]$, $p < .001$.

Prior task-specific AI use was also positively associated with automation preference, $b = 0.164$, $SE = 0.045$, 95\% CI $[0.075, 0.253]$, $p < .001$. Respondents who had previously used AI for a task therefore tended to prefer greater automation of that task after adjustment for the other modeled factors.

Concern that AI automation would make life less exciting or fulfilling was negatively associated with automation preference, $b = -0.077$, $SE = 0.022$, 95\% CI $[-0.120, -0.034]$, $p < .001$. Respondents who expressed greater concern about loss of fulfillment therefore tended to prefer less automation after adjustment for the other variables in the model.

The privacy composite was not independently associated with automation preference in the fully adjusted model, $b = 0.003$, $SE = 0.020$, 95\% CI $[-0.036, 0.043]$, $p = .862$. Desired collaboration was likewise not independently associated with automation, $b = 0.012$, $SE = 0.020$, 95\% CI $[-0.028, 0.051]$, $p = .554$.

Expected additional expenses were not statistically significant, $b = 0.031$, $SE = 0.022$, $p = .160$. Neither perceived identity consequences, $b = 0.012$, $SE = 0.019$, $p = .543$, nor consequences for the well-being of others, $b = -0.019$, $SE = 0.022$, $p = .388$, independently predicted automation preference.

None of the respondent-level technology-use covariates reached conventional statistical significance in the fully adjusted model. General attitude toward AI was positive in direction but nonsignificant, $b = 0.028$, $SE = 0.038$, $p = .468$. Income was also positive but did not reach the .05 threshold, $b = 0.039$, $SE = 0.021$, $p = .062$. Political orientation was likewise nonsignificant, $b = 0.024$, $SE = 0.021$, $p = .264$.

The model points to a relatively focused set of factors associated with delegation preferences. Participants favored greater automation when they trusted the AI more, wanted it to take a stronger leadership role, had previously used AI for the same task, or expected financial savings. Concern about losing fulfillment moved preferences in the opposite direction. Privacy and desired collaboration varied meaningfully across domains, but neither explained additional variation in automation preference once the other factors in the model were taken into account.

\subsubsection{Ordinal robustness analysis}

The cumulative ordinal GEE sensitivity analysis included all 2,601 observations from 289 respondents because its more parsimonious specification omitted respondent-level covariates containing missing values. The model retained respondent clustering, domain fixed effects, and all substantive task-level predictors, and produced the same substantive pattern for the central predictors.

Higher trust in AI was associated with substantially greater odds of selecting a higher automation category, $OR = 3.03$, 95\% CI $[2.52, 3.65]$, $p < .001$. Desired AI leadership was similarly associated with greater automation, $OR = 1.64$, 95\% CI $[1.41, 1.91]$, $p < .001$, as was expected financial savings, $OR = 1.36$, 95\% CI $[1.19, 1.55]$, $p < .001$.

Prior task-specific AI use was also associated with greater odds of selecting a higher automation category, $OR = 1.42$, 95\% CI $[1.17, 1.71]$, $p < .001$. Greater concern about loss of fulfillment was associated with lower odds of choosing a higher automation level, $OR = 0.80$, 95\% CI $[0.71, 0.91]$, $p < .001$.

As in the primary model, privacy concern was not independently associated with automation preference ($OR = 0.95$, 95\% CI $[0.87, 1.05]$, $p = .344$), nor was desired collaboration ($OR = 1.00$, 95\% CI $[0.91, 1.11]$, $p = .988$). Expected expenses, identity impact, and effects on others' well-being were likewise not significant in the ordinal sensitivity model. The consistency across specifications suggests that the main pattern of results was not sensitive to treating automation preference as approximately continuous in the primary analysis.

\subsubsection{Willingness to pay}

\begin{figure}[t]
    \centering
    \includegraphics[width=\linewidth]{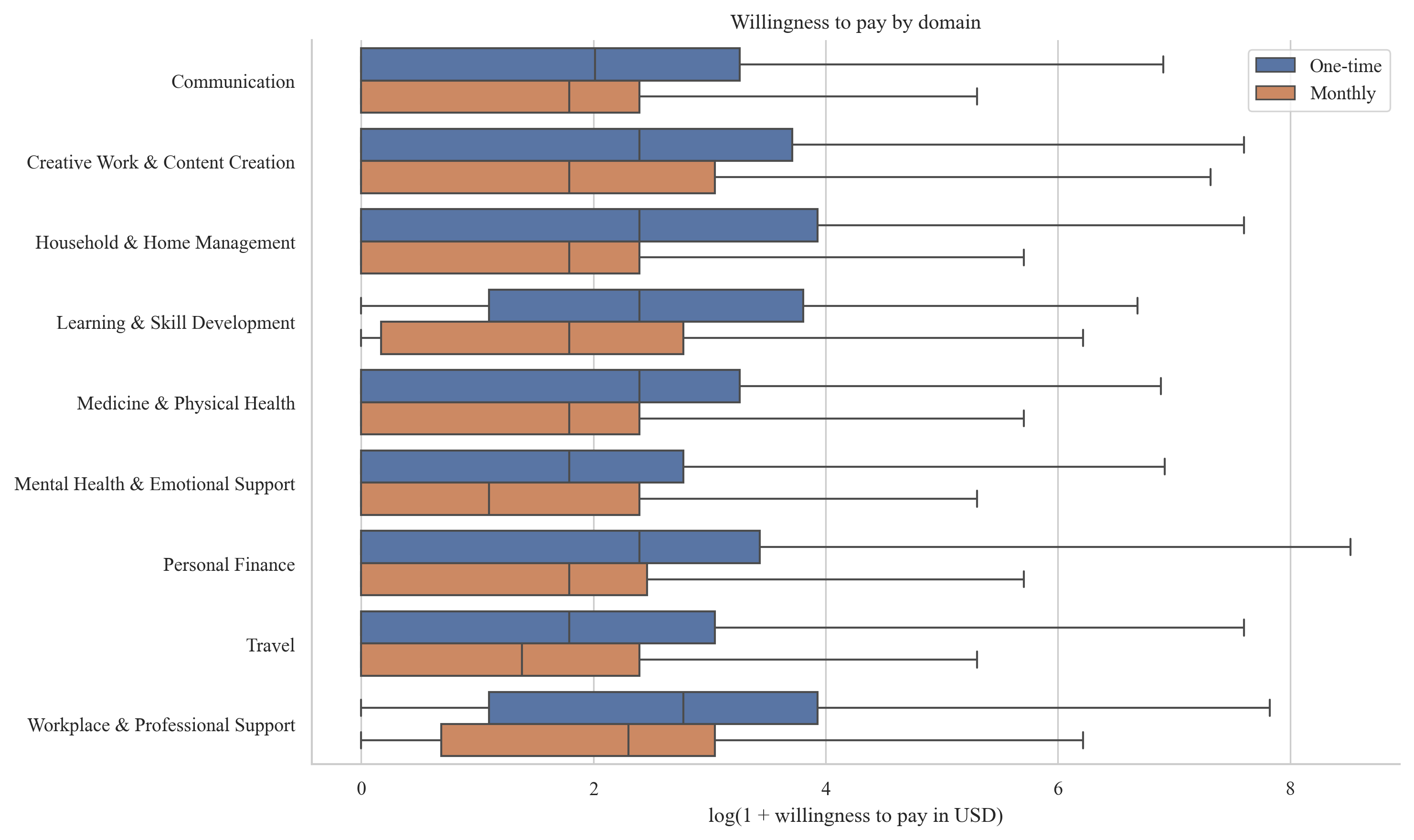}
    \caption{
        Willingness to pay for AI services across domains.
        Distributions are shown separately for one-time and monthly willingness to pay. Because monetary responses were highly right-skewed and included extreme values, the horizontal axis displays $\log(1+\mathrm{WTP})$. Outliers are omitted from the plotted boxplots for readability but were retained in the descriptive analyses.
    }
    \Description{
        Boxplots showing participants' willingness to pay for AI services across task domains, with separate distributions for one-time payments and monthly payments. The horizontal axis uses the transformed scale $\log(1+\mathrm{WTP})$ to accommodate highly right-skewed monetary responses. Outliers are not displayed in the figure but were retained in the descriptive analyses.
    }
    \label{fig:wtp}
\end{figure}

Willingness to pay (WTP) was generally modest, although responses were highly right-skewed and varied across domains (Figure~\ref{fig:wtp}). There were 2,429 usable one-time WTP responses. The median one-time WTP was \$10.00, with an interquartile range of \$0-\$30.00. Approximately 28.8\% of responses were \$0.  The upper tail was considerably larger: the 95th percentile was \$200 and the 99th percentile was approximately \$736, while the maximum reported value was \$200,000.

Monthly WTP followed a similar pattern. Among 2,426 valid responses, the median was \$5.00, with an interquartile range of \$0-\$15.00. Approximately 30.5\% of responses were \$0. The 95th percentile was \$60, the 99th percentile was \$250, and the maximum was \$5,000.

 The amount participants were willing to pay also differed by domain. Median one-time WTP was highest for Workplace \& Professional Support at \$15 and was \$10 for Creative Work \& Content Creation, Household \& Home Management, Learning \& Skill Development, Medicine \& Physical Health, and Personal Finance. Median one-time WTP was \$6.50 for Communication and \$5 for both Mental Health \& Emotional Support and Travel. Mental Health \& Emotional Support also had the highest share of zero one-time WTP responses, 43.9\%.

The monthly pattern was similar. Workplace \& Professional Support again had the highest median WTP at \$8.99. The median was \$5 for Communication, Creative Work \& Content Creation, Household \& Home Management, Learning \& Skill Development, Medicine \& Physical Health, and Personal Finance; it was \$3 for Travel and \$2 for Mental Health \& Emotional Support.

\subsubsection{Two-part willingness-to-pay models}

For one-time WTP, 2,429 complete observations were available for the positive-versus-zero model. Greater automation preference was associated with greater odds of being willing to pay a positive amount, $OR = 1.43$, $p < .001$, FDR-adjusted $p < .001$. Greater trust in AI was also associated with greater odds of positive WTP, $OR = 1.45$, $p < .001$, FDR-adjusted $p < .001$, as was greater expected savings, $OR = 1.41$, $p < .001$, FDR-adjusted $p < .001$. Privacy concern and expected additional expenses were not significantly associated with the likelihood of positive one-time WTP after correction.

Among the 1,729 observations with positive one-time WTP, greater expected savings predicted a larger payment amount, $b = 0.158$ on the log-WTP scale, $p < .001$, FDR-adjusted $p < .001$. Expected expenses were also positively associated with payment amount, $b = 0.107$, $p < .001$, FDR-adjusted $p < .001$. Privacy concern was positively associated with the amount respondents were willing to pay among positive responses, $b = 0.078$, $p = .007$, FDR-adjusted $p = .013$. Automation preference and AI trust were not significantly associated with the size of positive one-time payments after correction.

Monthly WTP showed a similar distinction. The positive-versus-zero model included 2,426 complete observations. Greater trust in AI predicted higher odds of reporting a positive monthly payment ($OR = 1.52$, $p < .001$, FDR-adjusted $p < .001$). Greater expected savings was also associated with positive monthly WTP, $OR = 1.39$, $p < .001$, FDR-adjusted $p < .001$, as was greater automation preference, $OR = 1.35$, $p < .001$, FDR-adjusted $p < .001$. Privacy concern and expected additional expenses were not significantly associated with positive monthly WTP after correction.

Among the 1,686 observations with positive monthly WTP, greater expected savings was associated with a larger payment amount, $b = 0.141$, $p < .001$, FDR-adjusted $p < .001$, as were greater expected expenses, $b = 0.127$, $p < .001$, FDR-adjusted $p < .001$. Privacy concern had a positive unadjusted association with monthly payment amount, $b = 0.051$, $p = .046$, but did not remain significant after FDR correction ($p = .077$). Neither automation preference nor AI trust was significantly associated with the size of positive monthly payments after correction. Across both payment formats, automation preference and trust were therefore more consistently related to whether participants would pay than to how much they would pay once they had assigned the service a positive monetary value.

\subsection{Qualitative Results}

\begin{figure*}[hbt]
  \centering
  \includegraphics[width=\textwidth]{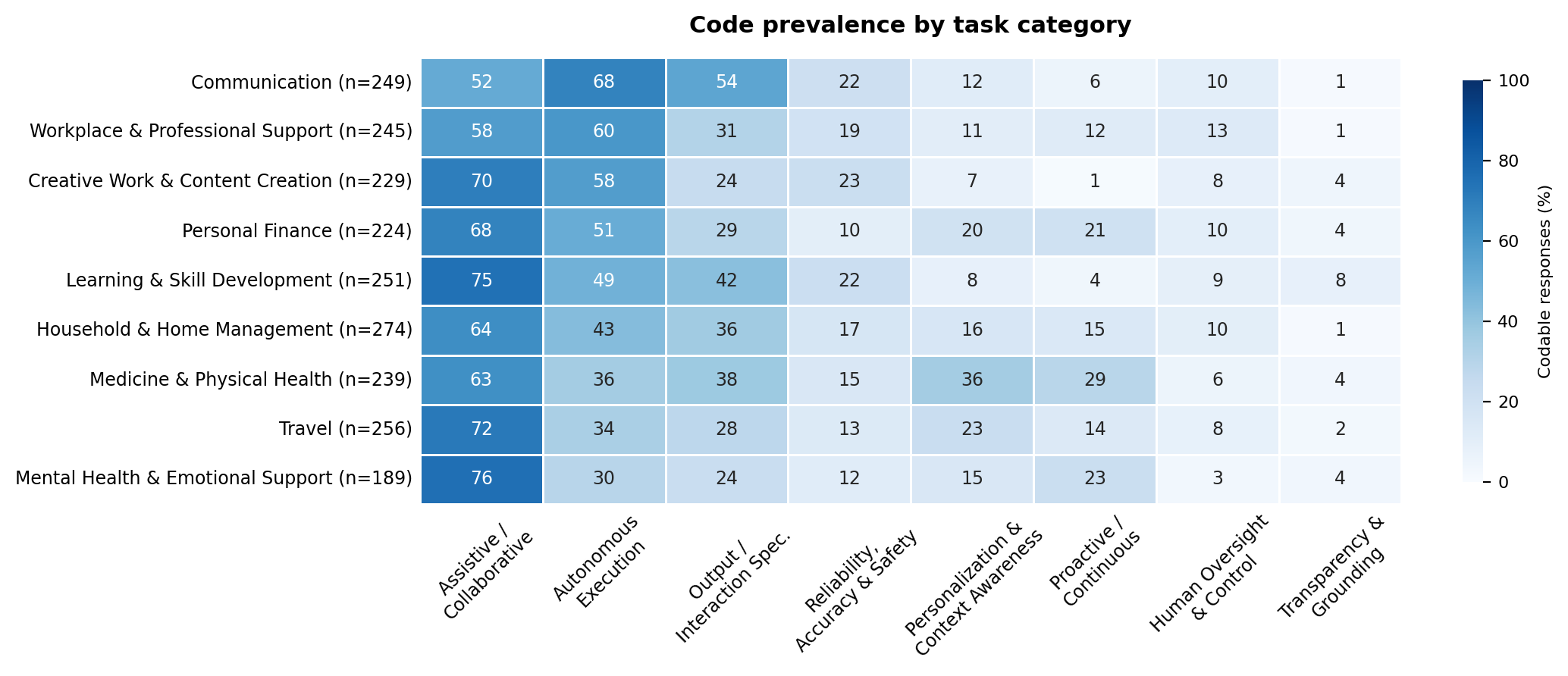}
  \caption{Code prevalence by task category. Cells show the percentage of codable open-ended responses within each domain carrying each code. Responses could receive multiple codes; percentages do not sum to 100\% within a row.}
  \Description{Heatmap showing the prevalence of qualitative codes across task categories. Rows represent task domains and columns represent qualitative codes. Each cell reports the percentage of codable open-ended responses in that domain assigned to the corresponding code, with darker or stronger shading indicating higher prevalence. Because responses could receive multiple codes, percentages within a row may sum to more than 100\%.}
  \label{fig:code-by-category}
\end{figure*}

The final corpus contained 2,216 non-empty responses, of which 2,156 were substantively codable. Multiple codes could be applied to each response when participants described more than one desired form of AI involvement.
The dominant pattern was that respondents rarely framed AI involvement as a simple choice between full automation and human performance. Figure~\ref{fig:code-by-category} presents the prevalence of each code across the nine task domains, revealing substantial variation in the forms of AI involvement participants described. Assistive/Collaborative Support appeared in 66.2\% of codable responses and Autonomous Execution in 48.0\%, with both appearing together in 25.6\%. Many responses described AI performing substantial portions of a task while leaving consequential decisions or final approval to the user. For example, one participant requested that an AI act “autonomously, but I want to approve and review major changes” (P292, Condition Adjustment). Similarly, for appointment scheduling, a participant wanted the AI to find and coordinate available times but to “ask for my approval before confirming any appointment” (P262). These responses illustrate a form of bounded autonomy in which task execution is delegated while judgment or final commitment remains under human control.

Desired configurations also varied meaningfully across tasks. Emotionally sensitive activities were framed primarily as collaborative as Assistive/Collaborative Support appeared in 91.9\% of \textit{Emotional Distress Support} responses and 90.0\% of \textit{Coping Strategies} responses. One participant described wanting AI to be “a sounding board and provide fresh ways to look at problems” (\textit{Emotional Distress Support}). By contrast, more procedural tasks more often elicited direct delegation, with Autonomous Execution appearing in 76.8\% of \textit{Appointment Scheduling} and 70.4\% of \textit{Translation} responses. Such contrasts suggest that users distinguish not simply how much AI should do, but which components of a task are appropriate to delegate.
Participants also frequently specified conditions surrounding delegation. Personalization and proactive assistance were particularly salient for activities that depend on changing user context. For language learning, one participant wanted an AI to “adapt to [their] level and track [their] progress”, while another wanted commute assistance to proactively incorporate “real-time traffic, weather, and [their] schedule”. Other responses emphasized accountability. In research tasks, participants requested citations, source links, and traceability; one participant asked for “page numbers and a hyperlink to exactly where in a PDF or a document it is drawing it from” (\textit{Research \& Academic Assistant}). For action-taking agents, transparency sometimes meant proof of execution rather than an agent's explanation of its results. For instance, the same participant wanted a subscription agent to cancel a service and simply “show [them] proof that it did.”

\subsection{Summary of findings}

Across the quantitative and qualitative analyses, preferences for AI delegation varied substantially across tasks and domains. Participants were generally more willing to automate activities such as travel and communication and less willing to automate mental-health and emotional-support tasks. In the adjusted models, greater trust in the AI agent, stronger preference for AI leadership, prior task-specific AI use, and anticipated financial savings were consistently associated with greater automation preference. Concern that automation would reduce personal fulfillment was associated with lower automation preference. These relationships were consistent across the Gaussian and ordinal model specifications. Privacy concern and desired human-AI collaboration varied across domains but were not independently associated with automation preference after adjustment.

The open-ended responses add detail to these patterns by showing how participants wanted delegation to work in practice. Many respondents combined substantial AI execution with continued human involvement, particularly through review, approval, or control over consequential decisions. The preferred arrangement also depended on the task. Emotionally sensitive activities elicited stronger preferences for collaborative support, whereas procedural activities more often elicited requests for autonomous execution. Participants further specified conditions for acceptable delegation, including personalization, proactive assistance, source traceability, verification of completed actions, and opportunities for oversight.

The findings distinguish several judgments that can be obscured by a single measure of automation preference. Participants separately considered how much of a task AI should perform, whether a human should remain involved, how much leadership the AI should have, and what forms of oversight or verification should accompany its actions. The following section considers what these distinctions mean for understanding delegation to AI agents and for designing systems that support different configurations of human and AI responsibility.

\section{Discussion}
Our findings suggest that preferences for AI delegation are more complex than a single continuum from human performance to full automation. Across the evaluated domains, participants varied substantially in how much of a task they wanted AI to perform, how much collaboration they desired, and how much leadership they were willing to give the AI. 
Trust in the AI, willingness to allow the AI to take a leading role, prior task-specific AI use, and anticipated financial savings were associated with greater automation preference, while concern that automation would reduce personal fulfillment was associated with lower automation preference.
Together with the qualitative findings, these results suggest that people make separate judgments about how much work AI should perform, how closely humans should remain involved, and who should retain decision authority.

\subsection{Task-Dependent Preferences for AI Delegation}

Our results reinforce prior work showing that willingness to rely on AI depends strongly on the task being performed. Previous studies have found lower acceptance of algorithmic systems for subjective or personally consequential activities and greater preferences for human involvement in such settings \cite{castelo2019task, longoni2019resistance}. Our findings extend this pattern across a broad set of everyday and work activities evaluated under a common survey structure. Participants showed greater willingness to automate communication and travel activities, while mental health, physical health, and professional tasks showed lower automation preferences. Trust in AI also varied across domains, with the lowest AI trust observed for Mental Health \& Emotional Support. These patterns are consistent with survey research showing that perceptions of AI risks and opportunities differ across application domains~\cite{schwesig2023using} and that context-specific evaluations can be more informative for use intentions than global attitudes toward AI~\cite{potinteu2023predicts}. Findings in this study further suggest that preferences for agentic AI are shaped not simply by general attitudes toward AI, but by the characteristics and consequences of the activity being delegated.

This task dependence also extends recent work on occupational automation. \citet{shao2025future} found considerable variation in workers' preferred levels of AI involvement across hundreds of occupational tasks, with workers often favoring human-AI partnership and retaining more human agency than technical experts considered necessary. Our study finds a related pattern outside paid work. This extension is important because much current generative-AI use occurs in non-work settings~\cite{chatterji2025how}, yet systematic comparisons of delegation preferences across everyday domains remain limited. By asking each participant to evaluate tasks across the same nine domains using a common set of measures, our design shows how delegation preferences change within individuals as the activity changes.
 
\subsection{Delegation, Collaboration, and Human Control}

Our findings suggest that AI delegation is better understood as a multidimensional configuration than as a single level of automation, which echoes the findings from previous studies \cite{lubars2019ask, bansal2021does,bucinca2021trust,sivaraman2023ignore}.
Desired AI leadership was independently associated with greater automation preference, while desired human–AI collaboration was not. This suggests that collaboration and automation should not be treated as opposing ends of the same continuum. A user may want an AI to perform substantial portions of a task while still wanting to remain involved in the process, and involvement does not necessarily imply that the human or AI must fully control the activity.

The qualitative findings provide further evidence of this distinction. Assistive or collaborative support appeared in 66.2\% of codable responses, autonomous execution appeared in 48.0\%, and both appeared together in 25.6\%. Participants frequently described workflows in which AI could complete substantial portions of an activity while human authority was preserved at important decision points. One participant wanted an AI to act “autonomously” while retaining the ability to “approve and review major changes,” while another wanted an agent to coordinate available appointments but request approval before confirming them. Emotionally sensitive tasks were more often described in collaborative terms, whereas procedural activities such as appointment scheduling and translation more frequently elicited autonomous execution. This distinction is compatible with prior evidence that people often prefer AI assistance over outright replacement of human professionals ~\cite{antes2021exploring,longoni2019resistance} and with observations that clinicians selectively accept, reject, or ``negotiate'' components of AI recommendations ~\cite{sivaraman2023ignore}. Together, these findings illustrate a recurring preference for bounded autonomy, allowing AI to independently execute substantial portions of a task while preserving human authority at consequential decision points.

Aforementioned findings also extend previous work on automation preference by suggesting that delegation itself may be measured too coarsely when represented only as how much of a task a user wants AI to perform. We conceptualize agentic delegation as involving several related but distinct dimensions: \textit{execution}, or how much of the task the AI performs; \textit{collaboration}, or how much the human remains involved in the process; and \textit{leadership}, or how much the AI directs the activity. Our qualitative findings further surface \textit{authority} as an important consideration: whether consequential actions remain subject to human approval, review, or override. These dimensions need not move together. Users may delegate execution without fully delegating authority, or remain extensively involved while allowing an AI to lead substantial portions of an activity. Agentic AI should therefore not be understood solely in terms of a level of automation, but also in terms of how execution, involvement, and decision authority are configured between human and AI actors. Future studies may benefit from measuring these dimensions separately rather than treating human involvement and automation as mutually exclusive.

\subsection{Trust, Fulfillment, and Design Implications}

Our findings suggest that decisions about AI delegation involve not only the benefits and risks of automation, but also the value of performing an activity oneself.
Aligned with previous works, we found that trust in AI was strongly associated with greater automation preference~\cite{lee2004trust} , while desired AI leadership, prior task-specific AI use, and anticipated financial savings were also consistently associated with greater automation~\cite{hoff2015trust}. By contrast, concern that automation would make life less exciting or fulfilling predicted lower automation preference. This finding is consistent with prior work showing that AI assistance can alter users’ sense of ownership, accomplishment, and engagement even when it improves task performance.

We characterize this consideration as the experiential value of human involvement, i.e., the value users derive from participating in the process of performing an activity, beyond the instrumental value of its outcome.
For some activities, the process of performing the task may itself have value. Learning, creativity, communication, and personal decision making may contribute to identity, enjoyment, or a sense of accomplishment in ways that are lost when the activity is fully delegated~\cite{draxler2024ghostwriter}. Consequently, increasing the technical capability of an agent does not necessarily imply that users will want greater automation. Systems may provide greater value when they reduce tedious or burdensome portions of an activity while preserving the aspects users want to perform themselves.

Technology-acceptance models have long emphasized perceived usefulness and effort expectancy~\cite{davis1989perceived,venkatesh2003user}, while consumer-oriented extensions explicitly incorporate price value as a tradeoff between perceived benefits and monetary costs~\cite{venkatesh2012consumer}. Our willingness-to-pay analyses further distinguish perceived value from payment magnitude. Greater automation preference, trust in AI, and expected savings were associated with whether respondents were willing to pay for an AI service at all, whereas expected savings and additional expenses were more consistently associated with the amount respondents were willing to pay among positive responses. This distinction suggests that willingness to adopt a paid agentic service and valuation of that service are related but separable decisions.

The findings also complicate a simple interpretation of privacy as a direct barrier to AI adoption. Privacy concern varied meaningfully across domains and was greatest in Mental Health \& Emotional Support and Personal Finance, followed by Medicine \& Physical Health and Workplace \& Professional Support. 
However, privacy concern was not independently associated with automation preference after accounting for trust, AI leadership, prior task-specific AI use, expected financial consequences, personal stakes, domain, and respondent characteristics.
Privacy therefore appears to operate as one consideration within a broader task-level evaluation rather than as a simple veto against AI use. More broadly, this finding suggests that delegation decisions concern not only what users risk by giving information to an AI, but also what they may give up by giving the activity itself to an AI. The former concerns informational exposure; the latter concerns the experiential value of continued human participation.

This pattern is particularly important for increasingly personalized agents. Participants frequently described wanting systems that could adapt to their preferences, track progress, monitor changing conditions, and proactively provide assistance. For example, one participant wanted a language-learning system to “adapt to my level and track my progress,” while another wanted commute assistance to incorporate “real-time traffic, weather, and my schedule.” These capabilities may improve usefulness while also requiring greater access to personal information. The resulting tension between personalization and privacy suggests that future agent designs should give users clearer control over what information is retained, how it is used, and which actions depend on it.

The qualitative findings also point to the importance of accountability when AI systems move from generating recommendations to taking actions. Participants requested citations, links, intermediate outputs, and confirmation that actions had been completed. In research tasks, respondents asked for source-level traceability, while in action-oriented tasks some wanted proof that the AI had successfully carried out the requested action. These findings suggest that transparency for agentic systems should include not only explanations of recommendations, but also visibility into what the system did, what information it used, and whether the action succeeded.

Overall, our findings suggest that the central design problem for agentic AI is not simply how much to automate, but how to configure delegation between humans and AI. Participants made distinct judgments about how much work an AI should execute, how closely humans should remain involved, how much leadership the AI should hold, and where consequential decision authority should remain. At the same time, these judgments reflected not only instrumental considerations such as trust and financial benefit, but also the experiential value users attached to participating in activities themselves. Together, these findings shift the design question from \textit{``How much should the AI automate?''} toward \textit{``What should be delegated, under whose authority, and what forms of human participation should be preserved?''} General-purpose agents may therefore benefit from configurable boundaries of autonomy that allow users to delegate execution while specifying approval checkpoints, retain meaningful participation in selected parts of an activity, and vary AI leadership across tasks or stages of the same task. Such designs would treat successful agentic AI not as the maximization of automation, but as the alignment of delegation with how users want work, authority, and meaningful participation to be distributed.

\subsection{Limitations and Future Work}
This study has several limitations. First, participants evaluated hypothetical daily life scenarios rather than interacting with functioning AI agents in these scenarios. Stated preferences may change when users experience real system errors, benefits, or interaction costs. Second, participants were relatively familiar with AI technologies: most reported substantial AI knowledge, high familiarity with LLMs, and frequent use of systems such as ChatGPT, Claude, and Gemini. Future work should examine whether similar patterns emerge among populations with less prior experience using AI who may reason about delegation differently.

Third, the study prioritized breadth across activities, meaning that respondents evaluated one randomly selected task within each domain rather than all possible tasks. The open-ended qualitative question was also presented only when respondents expressed at least some willingness to involve AI, so these responses primarily characterize how willing users want delegation to occur rather than why some users reject AI involvement altogether. Future studies could directly compare the preferences of users who accept and reject delegation and examine whether these preferences change through sustained interaction with AI agents in daily life tasks.

Finally, the present analysis examined average relationships across domains rather than testing whether the relationships between trust, leadership, privacy, fulfillment, and automation differ significantly from one domain to another. Future work could examine these interactions directly and investigate whether different forms of delegation emerge for particular categories of activity. Longitudinal studies would also help determine whether users maintain the same boundaries around AI authority after experiencing successful delegation, system failures, or changes in trust over time.

\section{Conclusion}

AI agents are increasingly capable of acting on users' behalf across a wide range of everyday and work activities, but willingness to delegate is highly task dependent. Through an online survey that collected responses from 289 U.S. adults across 45 scenarios spanning nine life domains, we found that automation preferences varied substantially by domain. Greater trust in the AI, desired AI leadership, prior task-specific AI use, and anticipated financial savings were associated with greater automation preference, whereas concern about losing personal fulfillment was associated with lower automation preference. Privacy concern also varied across domains, but it did not independently predict automation preference after adjustment.
Our findings further show that delegation is not well captured by a single continuum from human performance to full automation. Participants distinguished among how much of a task the AI should execute, how much collaboration they wanted, how much leadership the AI should hold, and where human authority should remain. Qualitative analysis on open-ended responses revealed a preference for bounded autonomy, in which AI could perform substantial portions of a task while humans retained oversight, approval, or final decision authority.
These results suggest that the design of agentic AI should focus not only on increasing capability, but also on configuring the boundaries of delegation. Systems that allow users to separately control execution, collaboration, leadership, and authority may better reflect how people want responsibility to be distributed between themselves and increasingly autonomous AI agents.




\bibliographystyle{ACM-Reference-Format}
\bibliography{references}

\appendix

\newcommand{\qq}[2]{%
  \par\medskip
  \noindent\textbf{#1}\enspace #2\par
}
\newcommand{\qopt}[1]{%
  \nopagebreak
  \par\noindent
  \hangindent=1.5em
  \textit{\small #1}\par
}
\newcommand{\qlogic}[1]{%
  \nopagebreak
  \par\noindent
  {\small\textsc{Display logic:} \textit{#1}}\par
}
\newcommand{\qtext}[1]{%
  \nopagebreak
  \par\noindent
  {\small\textit{(#1)}}\par
}


\section{Study Protocol and Survey Instrument}
\label{app:protocol}

Items are presented in their administered order. For readability, item labels in this document have been reassigned to reflect presentation order rather than Qualtrics' internal question IDs. Original Qualtrics IDs for the renumbered items are provided in the item-ID mapping table at the end of this document. Conditional display logic is stated immediately above the item it governs.

\subsection{Survey Flow}
\label{app:flow}

A screening branch terminated the survey for respondents who declined consent. Within each of the nine scenario blocks, the Qualtrics Loop \& Merge and Randomizer functionalities selected one of five candidate scenarios with equal probability, independently across blocks. All nine scenario blocks administered an identical battery in identical order.

\begin{table}[H]
\caption{Block structure of the administered survey.}
\label{tab:flow}
\small
\begin{tabular}{clc}
\toprule
\# & Block & Items \\
\midrule
1  & Consent form & 1 \\
2  & Demographic \& AI background & 14 \\
3  & Task instructions & 1 \\
4  & Scenarios --- Household \& Home Management & 23 \\
5  & Scenarios --- Workplace \& Professional Support & 23 \\
6  & Scenarios --- Travel & 23 \\
7  & Attention check 1 & 1 \\
8  & Scenarios --- Medicine \& Physical Health & 23 \\
9  & Scenarios --- Mental Health \& Emotional Support & 23 \\
10 & Scenarios --- Learning \& Skill Development & 23 \\
11 & Attention check 2 & 1 \\
12 & Scenarios --- Creative Work \& Content Creation & 23 \\
13 & Scenarios --- Communication & 23 \\
14 & Scenarios --- Personal Finance & 23 \\
15 & Compensation \& follow-up contact & 2 \\
16 & Completion redirect & 1 \\
\bottomrule
\end{tabular}
\end{table}

\subsection{Consent}
\label{app:consent}

\qq{C1.}{Welcome!}

You are being asked to participate in a voluntary research study. The purpose of this study is to understand how individuals integrate Artificial Intelligence (AI) agents into daily routines, and their attitudes toward expanded automation and augmentation by AI systems. Your participation is expected to last approximately thirty minutes. The risks associated with this research are minimal, and are no greater than those you ordinarily encounter in your daily activities. The benefits of participating in this research include contributing to a broader understanding of how AI technologies are used in everyday life and decision-making, while informing the development of AI agents that align with user needs and desires.

We invite you to take part in the research study because you are currently living in the United States, are older than 18 years of age, and have previously used AI agents. If you complete this research study, and your responses are deemed as legitimate (not generated by a Large Language Model or AI) by the research team, you will receive \$10 in the form of an Amazon gift card for your time and effort.

If you have questions, concerns, complaints, or think the research has harmed you, you may talk to the research team at \textsc{[redacted for review]}. This research is being overseen by an Institutional Review Board (``IRB''). You may also contact them at \textsc{[redacted for review]} if you have questions regarding your rights as a research participant or other questions, concerns, or complaints.

By clicking the button below, you acknowledge that your participation in the study is voluntary, you are 18 years of age, and that you are aware that you may choose to terminate your participation in the study at any time and for any reason.

Please note that this survey will be best displayed on a laptop or desktop computer. Some features may be less compatible for use on a mobile device.

\qopt{I consent, begin the study (1); I do not consent, I do not wish to participate (2)}
\qlogic{If option (2) is selected, the survey ends.}

\subsection{Demographic and AI Background}
\label{app:demographics}

\qlogic{Administered only to participants recruited through Prolific.}
\qq{D1.}{Please enter your unique Prolific ID.}
\qtext{open text}

\qq{D2.}{What is your age?}
\qopt{18--24 (1); 25--34 (2); 35--44 (3); 45--54 (4); 55--64 (5); 65--74 (6); 75--84 (7); 85 or older (8); Prefer not to say (9)}

\qq{D3.}{What is your gender?}
\qopt{Male (1); Female (2); Non-binary / third gender (3); Prefer not to say (4);
Other (5, open text)}

\qq{D4.}{What is your race/ethnicity?}
\qopt{White or Caucasian (1); Black or African American (2); American Indian/Native American or Alaska Native (3); Asian (4); Native Hawaiian or Other Pacific Islander (5); Other
(6, open text); Prefer not to say (7)}

\qq{D5.}{What was your total household income before taxes during the past 12 months?}
\qopt{Less than \$25,000 (1); \$25,000--\$49,999 (2); \$50,000--\$74,999 (3);
\$75,000--\$99,999 (4); \$100,000--\$149,999 (5); \$150,000 or more (6);
Prefer not to say (7)}

\qq{D6.}{What is the highest level of education you have completed?}
\qopt{Some high school or less (1); High school diploma or GED (2); Some college, but no degree (3); Associates or technical degree (4); Bachelor's degree (5); Graduate or professional degree (MA, MS, MBA, PhD, JD, MD, DDS etc.) (6); Prefer not to say (7)}

\qq{D7.}{What is your profession?}
\qtext{open text}

\qq{D8.}{What is your US Zip Code?}
\qtext{open text}

\qq{D9.}{In general, how would you describe your political orientation?}
\qopt{Very Liberal (1); Liberal (2); Slightly Liberal (3); Moderate (4); Slightly Conservative (5); Conservative (6); Very Conservative (7); Prefer not to say (8)}

\qq{D10.}{What is your overall attitude toward Artificial Intelligence (AI) technologies?}
\qopt{Very negative (1); Negative (2); Neutral (3); Positive (4); Very positive (5)}

\qq{D11.}{How knowledgeable would you say you are about Artificial Intelligence (AI)?}
\qopt{Not knowledgeable at all (1); Slightly knowledgeable (2); Moderately knowledgeable (3); Very knowledgeable (4); Extremely knowledgeable (5)}

\qq{D12.}{How familiar are you with Large Language Models (LLMs) such as ChatGPT, Claude, or Gemini?}
\qopt{Not familiar at all (1); Slightly familiar (2); Moderately familiar (3); Very familiar (4); Extremely familiar (5)}

\qq{D13.}{How often do you currently use LLMs (e.g., ChatGPT, Claude, Gemini) in your everyday life?}
\qopt{Never (1); Less than once a month (2); A few times a month (3); A few times a week (4); Daily (5); Multiple times a day (6)}

\qq{D14.}{For which types of tasks do you typically use AI or LLM-based tools in your own life? (Select all that apply)}
\qopt{Household Management (1); Travel (2); Professional Support (3); Medicine or Physical Health (4); Mental Health or Emotional Support (5); Learning or Skill Development (6); Content Creation (7); Communication (8); Personal Finances (9); Other (10, open text); I do not typically use AI tools (11, exclusive)}

\subsection{Task Instructions and AI Agent Definition}
\label{app:instructions}

\qq{I1.}{In the following portion of the survey, you will evaluate nine randomly-selected tasks. After reading each task, you will complete a set of questions assessing your preferences, comfort levels, perceived risks, and judgments about appropriate AI agent involvement in the task. Please note that this survey includes AI-generated images used to illustrate example scenarios. They do not depict real people or events.}

In this study, ``AI agent'' is defined as an Artificial Intelligence program capable of independently interpreting inputs, making decisions, and taking actions to achieve user-defined goals.

\subsection{Scenario Evaluation Battery}
\label{app:battery}

Each of the nine scenario blocks administered the same set of 22 response items and one instructional text item in the order shown below. This order was fixed across participants.

\paragraph{Scenario presentation.}
Task Summary: \verb|${lm://Field/2}| --- Task: \verb|${lm://Field/3}|, displayed with the accompanying AI-generated illustration. Field contents for all 45 scenarios are listed below.

\qq{1.}{Have you done this task before?}
\qopt{Yes (1); No (2)}

\qq{2.}{If you answered ``No'' to the task above, please answer all remaining questions
pertaining to this task hypothetically.}
\qtext{instructional text; displayed unconditionally}

\qlogic{Displayed if item 1 = Yes.}
\qq{3.}{How do you feel about performing this task?}
\qopt{Very negative (1); Negative (2); Neutral (3); Positive (4); Very positive (5)}

\qlogic{Displayed if item 1 = Yes.}
\qq{4.}{Have you used AI for this task before?}
\qopt{Yes (1); No (2)}

\qq{5.}{If an AI system could do this task for you completely, how much of the task would you want it to do?}
\qopt{None (1); A little (2); A moderate amount (3); A lot (4); The entire amount (5)}

\qlogic{Displayed if item 5 $\neq$ None.}
\qq{6.}{How would you like the AI system to perform the task for you?}
\qtext{open text}

\qq{7.}{To what extent do you anticipate saving money as a result of using AI to automate this task?}
\qopt{Not at all (1); A little (2); A moderate amount (3); A lot (4); A great deal (5)}

\qq{8.}{To what extent do you anticipate increased expenses as a result of using AI for this task?}
\qopt{Not at all (1); A little (2); A moderate amount (3); A lot (4); A great deal (5)}

\qq{9.}{If an AI could fully perform this task, how worried would you be that your life would become less exciting or fulfilling?}
\qopt{Not at all worried (1); Slightly worried (2); Moderately worried (3); Very worried (4); Extremely worried (5)}

\qq{10.}{If an AI system were to assist in this task, how much collaboration between you and the AI would be needed for successful completion?}
\qopt{No collaboration (1); Minimal collaboration (2); Moderate collaboration (3); Considerable collaboration (4); Essential collaboration (5)}

\qlogic{Displayed if item 5 = A lot or The entire amount.}
\qq{11.}{What are the main reasons you would want this task to be automated by AI?}
\begin{itemize}[nosep,leftmargin=1.5em]
\small
\item Automating this task would free up my time for higher-value activities. (1)
\item This task is repetitive or tedious. (2)
\item Automating this task would improve the quality of my life. (3)
\item This task is stressful or mentally draining. (4)
\item This task is complicated or difficult. (5)
\item Automating this task would help me stay organized when life becomes busy. (6)
\item Other (7, open text)
\end{itemize}

\qlogic{Displayed if item 5 = None or A little.}
\qq{12.}{What are the main reasons you would not want this task to be automated by AI?}
\begin{itemize}[nosep,leftmargin=1.5em]
\small
\item The consequences of a mistake would be too serious. (1)
\item I would want more human judgment or oversight. (2)
\item I would be concerned about how my personal data is used. (3)
\item I would not feel comfortable sharing the required information. (4)
\item This task requires human understanding, empathy, or emotional judgment. (5)
\item This task feels too personal to automate. (6)
\item I enjoy doing this task myself. (7)
\item I don't think automation would provide much benefit in this case. (8)
\item Other (9, open text)
\end{itemize}

\qq{13.}{How much do the results of this task affect your long-term well-being or identity?}
\qopt{Not at all (1); A little (2); A moderate amount (3); A lot (4); A great deal (5)}

\qq{14.}{How much does this task influence the wellbeing of the people you care about?}
\qopt{Not at all (1); A little (2); A moderate amount (3); A lot (4); A great deal (5)}

\qq{15.}{If an automated version of this task were offered as a one-time service, how much (in USD) would you be willing to pay?}
\qtext{open numeric text}

\qq{16.}{If an automated version of this task were offered as a monthly subscription service, how much (in USD) would you be willing to pay per month?}
\qtext{open numeric text}

\qq{17.}{How much do you trust an AI agent to complete this task for you?}
\qopt{Not at all (1); A little (2); A moderate amount (3); A lot (4); Completely (5)}

\qq{18.}{How much do you trust a human worker to complete this task for you?}
\qopt{Not at all (1); A little (2); A moderate amount (3); A lot (4); Completely (5)}

\qq{19.}{How much do you trust a human-AI team to complete this task?}
\qopt{Not at all (1); A little (2); A moderate amount (3); A lot (4); Completely (5)}

\qq{20.}{If a human-AI team were performing this task, how much would you want the AI to take a leading role?}
\qopt{Not at all (1); A little (2); A moderate amount (3); A lot (4); Entirely (5)}

\qq{21.}{To what extent do you think the AI agent involved in this scenario would be optimized for the platform provider's business goals rather than your personal needs?}
\qopt{Not at all (1); A little (2); A moderate amount (3); A lot (4); Entirely (5)}

\qq{22.}{In this scenario, how concerned are you about personal data being shared with third parties?}
\qopt{Not at all concerned (1); A little concerned (2); Moderately concerned (3); Very
concerned (4); Extremely concerned (5)}

\qq{23.}{In this scenario, how concerned are you about the AI system learning or inferring sensitive information about your preferences or personal profile?}
\qopt{Not at all concerned (1); A little concerned (2); Moderately concerned (3); Very concerned (4); Extremely concerned (5)}

\subsection{Attention Checks}
\label{app:attention}

Two instructed-response checks were embedded between scenario blocks, following the Travel block and the Learning \& Skill Development block respectively.

\qq{AC1.}{To ensure attention, please select the third option for this question.} 
\qopt{Strongly Disagree (1); Somewhat disagree (2); Neither agree nor disagree (3); Somewhat agree (4); Strongly agree (5)}

\qq{AC2.}{To ensure attention, please select the second option for this question.}
\qopt{Strongly disagree (1); Somewhat disagree (2); Neither agree nor disagree (3); Somewhat agree (4); Strongly agree (5)}

\subsection{Closing Items}
\label{app:closing}

\qlogic{Administered only to researcher-network participants.}
\qq{F1.}{Please provide the email address where you would like your Amazon e-gift card delivered.}
\qtext{open text}

\qq{F2.}{Would you be willing to be contacted for a follow-up interview at a later date?}
\qopt{Definitely not (1); Probably not (2); Might or might not (3); Probably yes (4); Definitely yes (5)}

\qlogic{Administered only to participants recruited through Prolific.}
\qq{F3.}{To show your completion, please click this link to be redirected back to [recruitment platform]: \textsc{[redacted for review]}}

\subsection{Scenario Descriptions}
\label{app:scenarios}

Tables~\ref{tab:scenario-text-a} and~\ref{tab:scenario-text-b} reproduce the task summary and task description exactly as shown to participants, for all 45 scenarios. The \# column gives the Loop \& Merge row index, which corresponds to the loop-iteration prefix on the exported response columns. Each scenario was additionally accompanied by an AI-generated illustration of the activity; the illustration files are omitted here.

\begin{table}[t]
\caption{Scenario descriptions as presented to participants, domains 1-5 of 9.}
\label{tab:scenario-text-a}
\small
\begin{tabular}{@{}rp{0.28\linewidth}p{0.58\linewidth}@{}}
\toprule
\# & Task summary & Description shown to participants \\
\midrule
\multicolumn{3}{@{}l}{\textbf{Household \& Home Management}} \\
1 & Recipe Search and Creation & Creating or finding recipes based on preferences and available ingredients. \\
2 & Home Systems Adjustment & Adjusting home lighting, temperature, and energy settings for comfort and cost efficiency. \\
3 & Home Improvement & Planning or figuring out how to complete home improvement or minor repair tasks. \\
4 & Home Safety Monitoring & Checking for signs of potential safety issues at home, such as gas leaks, fires, or intrusions. \\
5 & Entertainment Recommendation & Finding music, movies, or TV content aligned with your tastes. \\
\addlinespace
\multicolumn{3}{@{}l}{\textbf{Workplace \& Professional Support}} \\
6 & Reimbursement Processing & Organizing receipts, categorizing expenses, and preparing materials for reimbursement submissions. \\
7 & Job Application Support & Searching for jobs, filling out applications, and preparing for interviews. \\
8 & Hiring & Reviewing job applicants, evaluating resumes, and organizing candidates for hiring consideration. \\
9 & Threat Monitoring & Reviewing computer activity and system logs to identify potential security concerns. \\
10 & HR Communication & Drafting communication for HR tasks such as performance reviews or pay adjustment requests. \\
\addlinespace
\multicolumn{3}{@{}l}{\textbf{Travel}} \\
11 & Travel Booking & Finding and booking flights, hotels, and transit based on constraints and preferences. \\
12 & Lost Baggage Tracking & Following up on a lost or delayed bag by checking status updates, organizing information, and monitoring retrieval progress. \\
13 & Packing List Generation & Creating a tailored packing list based on your travel destination, season, and planned activities. \\
14 & Commute Optimization & Planning commutes by comparing routes, departure times, and transit options. \\
15 & Restaurant Recommendation & Finding restaurants based on dietary preferences and travel location. \\
\addlinespace
\multicolumn{3}{@{}l}{\textbf{Medicine \& Physical Health}} \\
16 & Symptom-Based Diagnostic Suggestions & Reviewing health symptoms (for yourself or others) and identifying possible explanations or next steps. \\
17 & Medication/Supplement Management & Managing information about supplements or medications and tracking dosage adherence. \\
18 & Health Monitoring & Keeping track of health-related information (e.g., daily steps, medications) over time to notice changes, patterns, or concerns. \\
19 & Condition Adjustment & Reflecting on health-related information to decide whether changes to daily routines, habits, or follow-up actions might be helpful. \\
20 & Physical Training Planning & Creating workout plans based on your goals, health history, and fitness data. \\
\addlinespace
\multicolumn{3}{@{}l}{\textbf{Mental Health \& Emotional Support}} \\
21 & Emotional Self-Check & Reflecting on mood, stress levels, or emotional well-being to identify patterns or areas of concern. \\
22 & Journaling & Writing about thoughts, feelings, or daily experiences to process emotions and gain personal insight. \\
23 & Coping Strategies & Identifying or exploring coping strategies for managing stress, anxiety, or everyday emotional challenges. \\
24 & Habit Formation & Tracking habits, setting reminders, and staying motivated toward personal or emotional well-being goals. \\
25 & Emotional Distress Support & Identifying ways to cope or feel more grounded during moments of intense emotional distress. \\
\bottomrule
\end{tabular}
\end{table}

\begin{table}[t]
\caption{Scenario descriptions as presented to participants, domains 6--9 of 9.}
\label{tab:scenario-text-b}
\small
\begin{tabular}{@{}rp{0.28\linewidth}p{0.58\linewidth}@{}}
\toprule
\# & Task summary & Description shown to participants \\
\midrule
\multicolumn{3}{@{}l}{\textbf{Learning \& Skill Development}} \\
26 & Language Learning & Studying and practicing a language by working on vocabulary, grammar, and conversational skills. \\
27 & Studying & Creating flashcards, quizzes, and mind maps from your study materials. \\
28 & Research \& Academic Assistant & Reviewing academic papers, summarizing key points, extracting citations, and outlining research topics. \\
29 & DIY/Hands-On Tasks & Finding step-by-step guidance for DIY hobby tasks. \\
30 & Coding & Writing, reviewing, and debugging code, and working through code examples. \\
\addlinespace
\multicolumn{3}{@{}l}{\textbf{Creative Work \& Content Creation}} \\
31 & Game Design & Coming up with ideas, implementing mechanics, or creating prototypes for personal game projects. \\
32 & Visual Design & Creating illustrations or design mockups from text-based descriptions. \\
33 & Video/Movie Creation \& Editing & Producing video scripts, selecting clips, and assembling edited videos. \\
34 & Writing & Drafting and revising written pieces, such as stories, articles, editorials, or newsletters. \\
35 & Research for Creative Content & Exploring books and related titles to develop ideas, themes, or concepts for creative projects. \\
\addlinespace
\multicolumn{3}{@{}l}{\textbf{Communication}} \\
36 & Communication Content Drafting & Drafting messages, emails, or daily chat responses. \\
37 & Translation & Translating uploaded materials or conversations between languages. \\
38 & Conversation Summarization & Summarizing or taking notes during meetings, calls, or long text conversations. \\
39 & Appointment Scheduling & Coordinating with providers to schedule appointments. \\
40 & Customer Service Support & Communicating with customer service to address questions, problems, or account issues. \\
\addlinespace
\multicolumn{3}{@{}l}{\textbf{Personal Finance}} \\
41 & Financial Report Analysis & Analyzing accounting records or financial statements and generating overviews. \\
42 & Account Aggregation & Bringing together information from your bank accounts, investments, and financial apps for analysis. \\
43 & Subscription Management & Identifying unused subscriptions and managing cancellations or renewals. \\
44 & Budget Allocation & Categorizing, creating, and tracking budget adjustments. \\
45 & Credit Management & Monitoring credit scores and payment schedules and determining possible methods for improvement. \\
\bottomrule
\end{tabular}
\end{table}

\subsection{Scenario Illustration Generation}
\label{app:illustrations}

Each of the 45 scenarios was accompanied by an AI-generated illustration intended to help participants visualize the activity being evaluated. The illustrations depicted the activity itself rather than a particular AI agent, interface, or embodiment.

The illustrations were generated using a common prompt structure with Gemini Nano Banana 2. Based on the retained generation prompts, prompts followed the form:

\begin{quote}
\textit{Generate a realistic image of the daily task of [task-specific activity description]. Do not include a person in the image.}
\end{quote}

Only the task-specific activity description varied across scenarios. Examples of retained prompts include:

\begin{itemize}
    \item \textbf{Recipe Search and Creation:}
    \textit{Generate a realistic image of the daily task of searching for and creating recipes based on available ingredients and dietary preferences. Do not include a person in the image.}

    \item \textbf{Home Systems Adjustment:}
    \textit{Generate a realistic image of the daily task of adjusting a home's lighting, temperature, and energy settings. Do not include a person in the image.}

    \item \textbf{Home Improvement Guidance:}
    \textit{Generate a realistic image of the daily task of following instructions for minor home repairs and improvements. Do not include a person in the image.}

    \item \textbf{Home Safety Monitoring:}
    \textit{Generate a realistic image of the daily task of monitoring a home for safety hazards such as gas leaks, fires, or intrusions. Do not include a person in the image.}

    \item \textbf{Entertainment Recommendation:}
    \textit{Generate a realistic image of the daily task of browsing and selecting music, movies, or TV content based on personal preferences. Do not include a person in the image.}
\end{itemize}

For each scenario, the first generated image that did not contain obvious visual defects or a substantial mismatch with the intended task was retained. Participants were informed that the illustrations were AI-generated and did not depict real people or events.

\subsection{Item-ID Mapping}
\label{app:id-mapping}

Table \ref{tab:item_mapping} maps the reader-facing item labels used in this supplementary document to the original Qualtrics question IDs. The 23-item scenario evaluation battery retains the sequential item numbering used throughout this document.

\begin{table}[H]
\centering
\caption{Mapping between supplementary item labels and original Qualtrics question IDs.}
\label{tab:item_mapping}
\small
\begin{tabular}{lll}
\toprule
Supplement label & Qualtrics ID & Item \\
\midrule
C1  & Q1   & Consent \\
D1  & Q240 & Prolific ID \\
D2  & Q9   & Age \\
D3  & Q10  & Gender \\
D4  & Q12  & Race/ethnicity \\
D5  & Q14  & Household income \\
D6  & Q15  & Education \\
D7  & Q239 & Profession \\
D8  & Q16  & ZIP code \\
D9  & Q17  & Political orientation \\
D10 & Q20  & Attitude toward AI \\
D11 & Q21  & AI knowledge \\
D12 & Q22  & LLM familiarity \\
D13 & Q23  & LLM use frequency \\
D14 & Q25  & AI/LLM task types \\
I1  & Q34  & Task instructions \\
F1  & Q225 & Gift-card email \\
F2  & Q236 & Follow-up interview interest \\
F3  & Q241 & Completion redirect \\
\bottomrule
\end{tabular}
\end{table}

\subsection{De-identification}
\label{app:deident}

Direct identifiers removed from the analytic datasets prior to analysis included IP addresses, gift-card email addresses, recruitment-platform identifiers, and geolocation information. Respondent identifiers retained for analysis were de-identified internal IDs.

\end{document}